\documentclass[twoside,leqno,twocolumn]{article}
\usepackage{ltexpprt}
\usepackage{subfigure}
\usepackage{graphicx}
\usepackage{amsmath}
\usepackage{bm}
\usepackage{url}
\usepackage{stmaryrd}
\usepackage{balance}
\usepackage{xcolor}

\usepackage{booktabs}

\usepackage{xcolor}
\usepackage{pgffor}
\usepackage{tikz}
\usetikzlibrary{shadows}
\usetikzlibrary{plotmarks}
\usetikzlibrary{fit}        
\usetikzlibrary{backgrounds}    
\usetikzlibrary{decorations.pathmorphing}
\usetikzlibrary{arrows,positioning}

\usetikzlibrary{decorations.pathmorphing} 
\usetikzlibrary{matrix} 
\usetikzlibrary{arrows} 
\usetikzlibrary{calc} 

\usepackage[usenames,dvipsnames]{pstricks}
\usepackage{epsfig}
\usepackage{pst-grad} 
\usepackage{pst-plot} 

\newcommand{\guc}{\textsc{Dug}}

\newcommand{\ie}[0]{\textit{i.e.}}
\newcommand{\eg}[0]{\textit{e.g.}}
\newcommand{\etc}[0]{\textit{etc.}}

\DeclareMathOperator*{\argmax}{arg\,max}

\newtheorem{defn}{\textsc{Definition}}

\newcommand{\frank}{\color{blue}}
\newcommand{\fbaseline}{\it \color{gray}}
\newcommand{\fexp}{\color{red}}
\newcommand{\fmed}{\color{green}}
\newcommand{\fmod}{\color{violet}}
\newcommand{\fphi}{\color{blue}}

\begin{document}
\title{Discriminative Feature Selection for Uncertain Graph Classification}
\author{
Xiangnan Kong\thanks{Department of Computer Science, University of Illinois at Chicago, USA. xkong4@uic.edu,  psyu@cs.uic.edu.}
\and
Philip S. Yu$^*$ \thanks{Computer Science Department, King Abdulaziz University, Jeddah, Saudi Arabia}
\and
Xue~Wang$^\mathcal{z}$
\and
Ann B. Ragin\thanks{Department of Radiology, Northwestern University, Chicago, IL, USA. \{xue-wang, ann-ragin\}@northwestern.edu.} 
}

\date{}
\maketitle
\begin{abstract}
Mining discriminative features for graph data has attracted much attention in recent years due to its important role in constructing graph classifiers, generating graph indices, etc. Most measurement of interestingness of discriminative subgraph features are defined on certain graphs, where the structure of graph objects are certain, and the binary edges within each graph represent the ``presence'' of linkages among the nodes. In many real-world applications, however, the linkage structure of the graphs is inherently uncertain. Therefore, existing measurements of interestingness based upon certain graphs are unable to capture the structural uncertainty in these applications effectively. In this paper, we study the problem of discriminative subgraph feature selection from uncertain graphs.  This problem is challenging and different from conventional subgraph mining problems because both the structure of the graph objects and the discrimination score of each subgraph feature are uncertain. To address these challenges, we propose a novel discriminative subgraph feature selection method, {\guc}, which can find discriminative subgraph features in uncertain graphs based upon different statistical measures including expectation, median, mode and $\varphi$-probability. We first compute the probability distribution of the discrimination scores for each subgraph feature based on dynamic programming. Then a branch-and-bound algorithm is proposed to search for discriminative subgraphs efficiently.  Extensive experiments on various neuroimaging applications ({\ie}, AlzheimerÕs Disease, ADHD and HIV) have been performed to analyze the gain in performance by taking into account structural uncertainties in identifying
discriminative subgraph features for graph classification. 
\end{abstract}


\section{Introduction}\label{sec:Introduction}
Graphs arise naturally in many scientific applications which involve complex structures in the data, {\eg}, chemical compounds, program flows, {\etc} Different from traditional data with flat features, these data are usually not directly represented as feature vectors, but as graphs with nodes and edges. Mining discriminative features for graph data has attracted much attention in recent years due to its important role in constructing graph classifiers, generating graph indices, etc. \cite{YCHY08,JYW10,CLZWY09,KY10,TCGH09}. Much of the past research in discriminative subgraph feature mining has focused on certain graphs, where the structure of the graph objects are certain, and the binary edges represent the ``presence'' of linkages between the nodes. Conventional subgraph mining methods \cite{YCHY08} utilize the structures of the certain graphs to find discriminative subgraph features. However, in many real-world applications, there is inherent uncertainty about the graph linkage structure.  Such uncertainty information will be lost if we directly transform uncertain graphs into certain graphs. 

\begin{figure}[t]
\centering
\subfigure[positive uncertain graph]{
    \begin{minipage}[l]{0.46\columnwidth}
      \centering
      \includegraphics[width=1\textwidth]{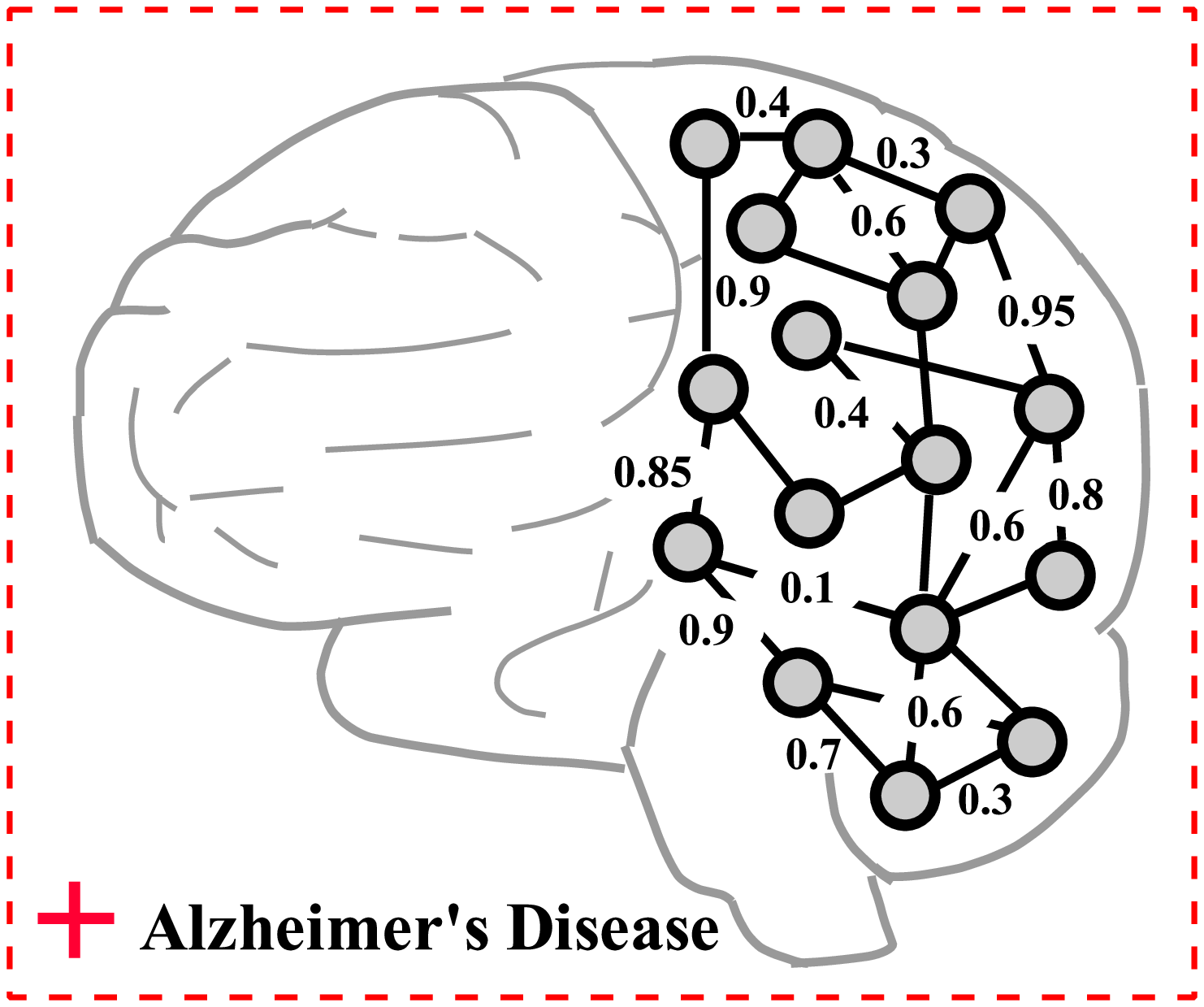}
    \end{minipage}
  }
  \subfigure[negative uncertain graph]{
    \begin{minipage}[l]{0.46\columnwidth}
      \centering
      \includegraphics[width=1\textwidth]{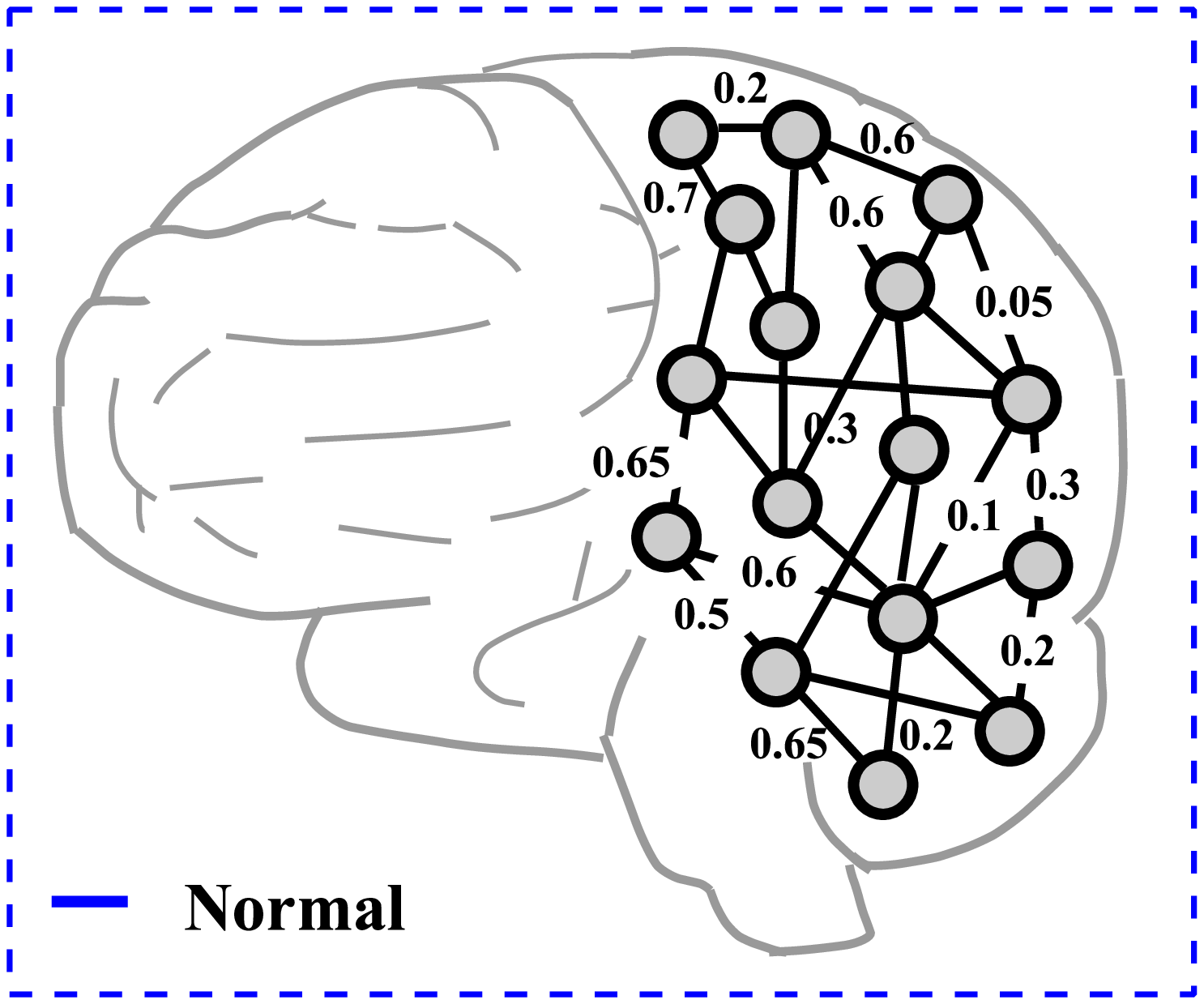}
    \end{minipage}
  }
\caption{An example of uncertain graph classification task. } \label{fig:eg_brain}
\end{figure}

For example, in neuroimaging, the functional connectivities among different brain regions are highly uncertain  \cite{HLS09,HSY10,HLY11,ZYLY11}. In such applications, each human brain can be represented as an uncertain graph as shown in Figure~\ref{fig:eg_brain}, which is also called the ``brain network" \cite{BS09}.  In such brain networks, the nodes represent brain regions, and edges represent the probabilistic connections, {\eg}, resting-state functional connectivity in fMRI (functional Magnetic Resonance Imaging).  Since these functional connectivities are derived based upon processing steps, such as temporal correlations in spontaneous blood oxygen level-dependent (BOLD) signal oscillations, each edge of the brain network is associated with a probability to quantify the likelihood that the functional connection exists in the brain. Resting-state functional connectivity has shown alterations related to many neurological diseases, such as ADHD (Attention Deficit Hyperactivity Disorder), Alzheimer's disease and  virus infections that may affect the brain functioning, such as HIV \cite{WFOC11}. Researchers are interested in analyzing the complex structure and uncertain connectivities of the human brain to find biomarkers for neurological diseases. Such biomarkers are clinically imperative for detecting injury to the brain in the earliest stages before it is irreversible. Valid biomarkers can be used to aid diagnosis, monitor disease progression and evaluate effects of intervention.



Motivated by these real-world neuroimaging applications, in this paper, we study the problem of mining discriminative subgraph features in uncertain graph datasets. Discriminative subgraph features are fundamental for uncertain graphs, just as they are for certain graphs. They serve as primitive features for the classification tasks on uncertain graph objects. Despite the value and significance, the discriminative subgraph mining for uncertain graph classification has not been studied in this context. If we consider discriminative subgraph mining and uncertain graph structures as a whole, the major research challenges are as follows:

\noindent\textbf{Structural Uncertainty:} In discriminative subgraph mining, we need to estimate the discrimination score of a subgraph feature in order to select a set of subgraphs that are most discriminative for a classification task. In conventional subgraph mining, the discrimination scores of subgraph features are defined on certain graphs, where the structure of each graph object is certain, and thus the containment relationships between subgraph features and graph objects are also certain. However, when uncertainty is presented in the structures of graphs, a subgraph feature only exists within a graph object with a probability. Thus the discrimination scores of a subgraph feature are no longer deterministic values, but random variables with probability distributions.

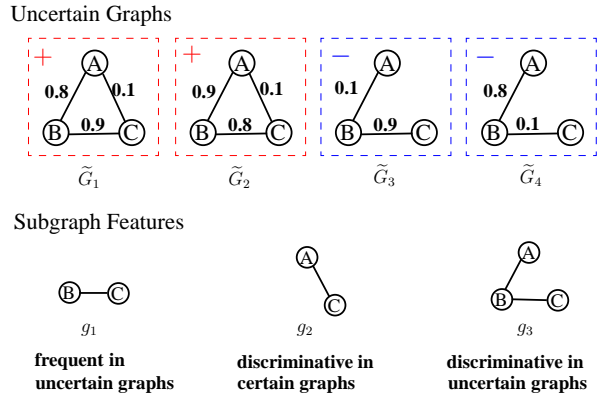
\begin{figure}[t]
\centering
\scalebox{0.6} 
{
\begin{pspicture}(1,-4.4)(15.1,4.4)
\pscircle[linewidth=0.04,dimen=outer](2.8318555,3.1130857){0.3}
\usefont{T1}{ptm}{m}{n}
\rput(2.8306055,3.1130857){\Large A}
\pscircle[linewidth=0.04,dimen=outer](1.9718554,1.5730857){0.3}
\usefont{T1}{ptm}{m}{n}
\rput(1.9548241,1.5730857){\Large B}
\pscircle[linewidth=0.04,dimen=outer](3.6518555,1.5730857){0.3}
\usefont{T1}{ptm}{m}{n}
\rput(3.6349413,1.5730857){\Large C}
\psline[linewidth=0.04cm](2.6318555,2.913086)(2.0718555,1.8330857)
\psline[linewidth=0.04cm](3.3918555,1.5530857)(2.2518554,1.5330856)
\pscircle[linewidth=0.04,dimen=outer](12.551855,3.1130857){0.3}
\usefont{T1}{ptm}{m}{n}
\rput(12.536856,3.1180856){\Large A}
\pscircle[linewidth=0.04,dimen=outer](11.691855,1.5730857){0.3}
\usefont{T1}{ptm}{m}{n}
\rput(11.674824,1.5730857){\Large B}
\pscircle[linewidth=0.04,dimen=outer](13.371856,1.5730857){0.3}
\usefont{T1}{ptm}{m}{n}
\rput(13.354941,1.5730857){\Large C}
\psline[linewidth=0.04cm](12.351855,2.913086)(11.791856,1.8330857)
\psline[linewidth=0.04cm](13.1118555,1.5530857)(11.971855,1.5330856)
\psframe[linewidth=0.02,linecolor=red,linestyle=dashed,dash=0.16cm 0.16cm,dimen=outer](4.2518554,3.673086)(1.3518555,1.0530857)
\usefont{T1}{ptm}{b}{n}
\rput(1.989121,2.4830856){\large 0.8}
\usefont{T1}{ptm}{b}{n}
\rput(2.7886913,1.7430856){\large 0.9}
\usefont{T1}{ptm}{b}{n}
\rput(11.709121,2.5030856){\large 0.8}
\usefont{T1}{ptm}{b}{n}
\rput(12.432597,1.7630856){\large 0.1}
\usefont{T1}{ptm}{b}{n}
\rput(1.6857617,3.2499998){\LARGE \color{red}$\mathbf{+}$}
\usefont{T1}{ptm}{b}{n}
\rput(11.485762,3.2499998){\LARGE \color{blue}$\mathbf{-}$}
\psframe[linewidth=0.02,linecolor=blue,linestyle=dashed,dash=0.16cm 0.16cm,dimen=outer](13.951856,3.673086)(11.051855,1.0530857)
\usefont{T1}{ptm}{b}{n}
\rput(2.709668,0.58308566){\large $\widetilde{G}_1$}
\pscircle[linewidth=0.04,dimen=outer](6.0918555,3.1130857){0.3}
\usefont{T1}{ptm}{m}{n}
\rput(6.0906053,3.1130857){\Large A}
\pscircle[linewidth=0.04,dimen=outer](5.2318554,1.5730857){0.3}
\usefont{T1}{ptm}{m}{n}
\rput(5.214824,1.5730857){\Large B}
\pscircle[linewidth=0.04,dimen=outer](6.9118557,1.5730857){0.3}
\usefont{T1}{ptm}{m}{n}
\rput(6.8949413,1.5730857){\Large C}
\psline[linewidth=0.04cm](5.8918552,2.913086)(5.3318553,1.8330857)
\psline[linewidth=0.04cm](6.6518555,1.5530857)(5.5118556,1.5330856)
\psframe[linewidth=0.02,linecolor=red,linestyle=dashed,dash=0.16cm 0.16cm,dimen=outer](7.5118556,3.673086)(4.6118555,1.0530857)
\usefont{T1}{ptm}{b}{n}
\rput(5.2486916,2.4830856){\large 0.9}
\usefont{T1}{ptm}{b}{n}
\rput(6.049121,1.7430856){\large 0.8}
\usefont{T1}{ptm}{b}{n}
\rput(4.9857616,3.308086){\LARGE \color{red}$\mathbf{+}$}
\pscircle[linewidth=0.04,dimen=outer](9.311855,3.1130857){0.3}
\usefont{T1}{ptm}{m}{n}
\rput(9.310605,3.1130857){\Large A}
\pscircle[linewidth=0.04,dimen=outer](8.451856,1.5730857){0.3}
\usefont{T1}{ptm}{m}{n}
\rput(8.434824,1.5730857){\Large B}
\pscircle[linewidth=0.04,dimen=outer](10.131855,1.5730857){0.3}
\usefont{T1}{ptm}{m}{n}
\rput(10.114942,1.5730857){\Large C}
\psline[linewidth=0.04cm](9.1118555,2.913086)(8.551855,1.8330857)
\psline[linewidth=0.04cm](9.871856,1.5530857)(8.731855,1.5330856)
\psframe[linewidth=0.02,linecolor=blue,linestyle=dashed,dash=0.16cm 0.16cm,dimen=outer](10.731855,3.673086)(7.8318553,1.0530857)
\usefont{T1}{ptm}{b}{n}
\rput(8.412598,2.5830855){\large 0.1}
\usefont{T1}{ptm}{b}{n}
\rput(9.268691,1.7430856){\large 0.9}
\usefont{T1}{ptm}{b}{n}
\rput(8.285762,3.2680857){\LARGE \color{blue}$\mathbf{-}$}
\usefont{T1}{ptm}{b}{n}
\rput(6.035791,0.58308566){\large $\widetilde{G}_2$}
\usefont{T1}{ptm}{b}{n}
\rput(9.253858,0.6230856){\large $\widetilde{G}_3$}
\usefont{T1}{ptm}{b}{n}
\rput(12.497617,0.6230856){\large $\widetilde{G}_4$}
\psline[linewidth=0.04cm](3.5118554,1.7930856)(3.0118554,2.8730857)
\usefont{T1}{ptm}{b}{n}
\rput(3.4725978,2.5230854){\large 0.1}
\usefont{T1}{ptm}{b}{n}
\rput(6.7525973,2.5430853){\large 0.1}
\psline[linewidth=0.04cm](6.7518554,1.8130857)(6.2518554,2.8930857)
\usefont{T1}{ptm}{m}{n}
\rput(2.751953,4.168086){\Large Uncertain Graphs}
\usefont{T1}{ptm}{m}{n}
\rput(2.9231834,-0.45640653){\Large Subgraph Features}

\pscircle[linewidth=0.04,dimen=outer](2.2750394,-1.9741606){0.25}
\usefont{T1}{ptm}{m}{n}
\rput(2.2560942,-1.9741606){\large B}
\pscircle[linewidth=0.04,dimen=outer](3.355039,-1.9941605){0.25}
\usefont{T1}{ptm}{m}{n}
\rput(3.3362012,-1.9941605){\large C}
\psline[linewidth=0.04cm](3.1068554,-1.9836329)(2.4868555,-1.9836329)

\pscircle[linewidth=0.04,dimen=outer](7.5350385,-1.1941603){0.25}
\usefont{T1}{ptm}{m}{n}
\rput(7.529726,-1.1941603){\large A}
\pscircle[linewidth=0.04,dimen=outer](8.15504,-2.2541604){0.25}
\usefont{T1}{ptm}{m}{n}
\rput(8.136202,-2.2541604){\large C}
\psline[linewidth=0.04cm](7.6668553,-1.3836329)(8.026855,-2.063633)

\usefont{T1}{ptm}{b}{n}\rput(2.7118554,-2.8){\large $g_1$}
\usefont{T1}{ptm}{b}{n}\rput(7.4918556,-2.8){\large $g_2$}
\usefont{T1}{ptm}{b}{n}\rput(12.371855,-2.8){\large $g_3$}

\usefont{T1}{ptm}{m}{n}\rput(2.5,-3.5){\large \textbf{frequent in}}
\usefont{T1}{ptm}{m}{n}\rput(3,-4){\large \textbf{ uncertain graphs}}
\usefont{T1}{ptm}{m}{n}\rput(7.5,-3.5){\large \textbf{discriminative in}}
\usefont{T1}{ptm}{m}{n}\rput(7.3,-4){\large \textbf{certain graphs}}
\usefont{T1}{ptm}{m}{n}\rput(12.2,-3.5){\large \textbf{discriminative in }}
\usefont{T1}{ptm}{m}{n}\rput(12.2,-4){\large \textbf{uncertain graphs}}

\pscircle[linewidth=0.04,dimen=outer](12.451856,-1.0869144){0.25}
\usefont{T1}{ptm}{m}{n}
\rput(12.446543,-1.0869144){\large A}
\pscircle[linewidth=0.04,dimen=outer](11.851855,-2.1269143){0.25}
\usefont{T1}{ptm}{m}{n}
\rput(11.832911,-2.1269143){\large B}
\pscircle[linewidth=0.04,dimen=outer](13.091855,-2.1469142){0.25}
\usefont{T1}{ptm}{m}{n}
\rput(13.073017,-2.1469142){\large C}
\psline[linewidth=0.04cm](12.286856,-1.223633)(11.926855,-1.903633)
\psline[linewidth=0.04cm](12.846855,-2.143633)(12.066855,-2.1236331)
\end{pspicture} 
}
\caption{Different types of subgraph features for uncertain graph classification} \label{fig:subproperty}\vspace{-10pt}
\end{figure}

Thus, the  evaluation of discrimination scores for subgraph features in uncertain graphs is different from conventional subgraph mining problems. For example, in Figure~\ref{fig:subproperty}, we show an uncertain graph dataset containing 4 uncertain graphs $\widetilde{G}_1, \cdots, \widetilde{G}_4$ with their class labels, $+$ or $-$. Subgraph $g_1$ is a frequent pattern among the uncertain graphs, but it may not relate to the class labels of the graphs.  Subgraph $g_2$ is a discriminative subgraph features when we ignore the edge uncertainties. However, if such uncertainties are considered, we will find that $g_2$ can rarely be observed within the uncertain graph dataset, and thus will not be useful in graph classification.  Accordingly, $g_3$ is the best subgraph feature for uncertain graph classification.

\noindent\textbf{Efficiency \& Robustness:} There are two additional  problems that need to be considered when evaluating features for uncertain graphs: 1) In an uncertain graph dataset, there are an exponentially large number of possible instantiations of a graph dataset \cite{ZGL10}. How can we efficiently compute the discrimination score of a subgraph feature without enumerating  all possible implied datasets? 2) When evaluating the subgraph features, we should choose a statistical measure for the probablity disctribution of discrimination scores which is robust to extreme values. For example,  given  a subgraph feature with (score, probability) pairs as $(0.01, 99.99\%)$ and $(+\infty, 0.01\%)$, the \emph{expected} score of the subgraph is $+\infty$, although this value is only associated with a very tiny probability.





In order to address the above problems, we propose a general framework for mining discriminative subgraph features in uncertain graph datasets, which is  called {\guc} (Discriminative feature selection for Uncertain Graph classification). The {\guc} framework can effectively find a set of discriminative subgraph features by considering the relationship between uncertain graph structures and  labels based upon various statistical measures.  We propose an efficient method to calculate the probability distribution of the scoring function based on dynamic programming. Then a branch-and-bound algorithm is proposed to search for the discriminative subgraphs efficiently by pruning the subgraph search space.  Empirical studies on resting-state fMRI images of different brain diseases (i.e., Alzheimer's Disease, ADHD and HIV) demonstrate that the proposed method can obtain better accuracy on uncertain graph classification tasks than alternative approaches. 

For the rest of the paper, we first introduce preliminaries in Section~\ref{sec:prob_form}. Then we introduce our {\guc} subgraph mining framework in Section~\ref{subsec:guc}.  Discrimination score functions based upon different statistic measures are discussed in Section~\ref{subsec:score_dist}. An efficient algorithm for computing the score distribution based upon dynamic programming is proposed in Section~\ref{subsec:dist_comp}. Experimental results are discussed in Section~\ref{sec:experiment}. In Section~\ref{sec:conclusion}, we conclude the paper.

\begin{table*}[t]
    \centering
    \caption{Important Notations.}\label{tab:notation}
    {\scriptsize
    \begin{tabular}{ll}
    \toprule
    Symbol&  Definition\\
    \midrule
    $\widetilde{\mathcal{D}}=\{\widetilde{G}_1, \cdots, \widetilde{G}_n\}$ & uncertain graph dataset, $\widetilde{G}_i$ denotes the $i$-th uncertain graph in the dataset.\\
    $\mathbf{y}=[y_1, \cdots, y_n]^\top$    &  class label vector for graphs in $\widetilde{\mathcal{D}}$, $y_i\in \{+1,-1\}$.\\
    $\widetilde{\mathcal{D}}_+ $ and $\widetilde{\mathcal{D}}_- $   & the subset of $\widetilde{\mathcal{D}}$ with only positive/negative graphs, $\widetilde{\mathcal{D}}_+ =\{\widetilde{G}_i|\widetilde{G}_i \in \widetilde{\mathcal{D}}, y_i =+1\}$.\\
    $n_+ $ and $n_-$ & number of positive graphs and number of negative graphs in $\widetilde{\mathcal{D}}$, $n_+ =|\widetilde{\mathcal{D}}_+|$ and $n_-=|\widetilde{\mathcal{D}}_-|$.\\
	$\mathcal{D} = \{G_1, \cdots, G_n\}$ & a certain graph dataset implied from $\widetilde{\mathcal{D}}$, $G_i$ denotes the certain graph implied from $\widetilde{G}_i$.\\
	$g\subseteq G$  & graph $G$ contains subgraph feature $g$\\
	$n_+^g$ and $n_-^g$ & number of graphs in $\mathcal{D}_+$ / $\mathcal{D}_-$ that contains subgraph $g$,  $n_+^g=|\{G_i | g\subseteq G_i, G_i\in \mathcal{D}_+\}|$.\\
	$\widetilde{G}\Rightarrow G$ and $\widetilde{\mathcal{D}}\Rightarrow \mathcal{D}$& certain graph $G$ is implied from uncertain graph $\widetilde{G}$; $\mathcal{D}$ is implied from $\widetilde{\mathcal{D}}$. \\
    $\mathcal{W}(\widetilde{G}) $ and $\mathcal{W}(\widetilde{\mathcal{D}})$ & the possible worlds of $\widetilde{G}$ and  $\widetilde{\mathcal{D}}$, $\mathcal{W}(\widetilde{G}) = \{G| \widetilde{G}\Rightarrow G\}$, $\mathcal{W}(\widetilde{\mathcal{D}}) = \{ \mathcal{D} | \widetilde{\mathcal{D}}\Rightarrow \mathcal{D}\}$.\\
    $E(\widetilde{G}_i)$ and  $E(G_i)$ & the set of edges in $\widetilde{G}_i$ and $G_i$ \\
    $\widetilde{\mathcal{D}}_+(k)$ and  $\widetilde{\mathcal{D}}_-(k)$ & the first $k$ graphs in $\widetilde{\mathcal{D}}_+$ or $\widetilde{\mathcal{D}}_-$\\
     \bottomrule
    \end{tabular}
    }
\end{table*}

\section{Problem Formulation}\label{sec:prob_form}

In this section, we formally define the model of uncertain graphs and the problem of discriminative subgraph mining in uncertain graph datasets.  
Suppose we are given an uncertain graph dataset $\widetilde{\mathcal{D}}= \{ \widetilde{G}_1, \cdots, \widetilde{G}_{n}\}$ that consists of $n$ uncertain graphs. $\widetilde{G}_{i}$ is the $i$-th uncertain graph in $\widetilde{\mathcal{D}}$. $\mathbf{y}=[y_1, \cdots, y_n]^{\top}$ denotes the vector of class labels, where $y_i\in\{+1, -1\}$ is the class label of $\widetilde{G}_i$.  We also denote the subset of $\widetilde{\mathcal{D}}$ that contains only positive/negative graphs as $\widetilde{\mathcal{D}}_+ = \{\widetilde{G}_i  | \widetilde{G}_i \in \widetilde{\mathcal{D}} \bigwedge y_i = +1\}$ and $\widetilde{\mathcal{D}}_- = \{\widetilde{G}_i  | \widetilde{G}_i \in \widetilde{\mathcal{D}} \bigwedge  y_i=-1\}$ respectively.

\begin{defn}[Certain Graph]
A certain graph is an undirected and deterministic graph represented as $G = (V, E)$.
$V=\{ v_1, \cdots, v_{n_v} \}$ is the set of vertices. $E\subseteq V \times V$ is the set of deterministic edges. 
\end{defn}

\begin{defn}[Uncertain Graph]
An uncertain graph is an undirected and nondeterministic graph represented as $\widetilde{G} = (V, E, p)$.
$V=\{ v_1, \cdots, v_{n_v} \}$ is the set of vertices.  $E\subseteq V \times V$ is the set of nondeterministic edges. $p: E\rightarrow (0,1]$  is a function that assigns a probability of existence to each edge in $E$.  $p(e)$ denotes the existence probability of edge $e\in E$.
\end{defn}

Consider an uncertain graph $\widetilde{G}(V, E, p) \in \widetilde{\mathcal{D}}$, where each edge $e\in E$ is associated with a probability $p(e)$ of being present.  As in previous works \cite{ZLGZ09,ZGL10},  we assume that the uncertainty variables of different edges in an uncertain graph are independent from each other,  though most of our results are still applicable to graphs with edge correlations. We further assume that all uncertain graphs in a dataset $\widetilde{\mathcal{D}}$ share a same set of nodes $V$ and each node in $V$ has a unique node label, which is reasonable in many applications like neuroimaging, since each human brain consists of the same number of regions. The main difference between different uncertain graphs is on their linkage structures, {\ie}, the edge sets $E(\widetilde{G})$ and the edge probabilities $p(e)$.

Possible instantiations of an uncertain graph $\widetilde{G}$ are usually referred to as \emph{worlds} of $\widetilde{G}$, where each world corresponds to an implied certain graph $G$.   Here $G$ is \emph{implied} from uncertain graph $\widetilde{G}$ (denoted as $\widetilde{G}\Rightarrow G$), iff all edges in $E(G)$ are sampled from  $E(\widetilde{G})$ according to their probabilities of existence in $p(e)$ and $E(G)\subseteq E(\widetilde{G})$. There are $2^{|E(\widetilde{G})|}$ possible worlds for uncertain graph $\widetilde{G}$, denoted as $\mathcal{W}(\widetilde{G}) = \{G | \widetilde{G} \Rightarrow G\}$. Thus, each uncertain graph $\widetilde{G}$ corresponds to a probability distribution over $\mathcal{W}(\widetilde{G})$. We denote the probability of each certain graph $G\in \mathcal{W}(\widetilde{G})$ being implied by the uncertain graph $\widetilde{G}$ as $\Pr(\widetilde{G} \Rightarrow G)$, and we have
{
$$ \Pr\!\left[\widetilde{G} \Rightarrow G\right] = \prod_{e\in E(G)} \Pr\!_{\widetilde{G}}(e) \prod_{e\in E(\widetilde{G})-E(G)} \left(\ 1-\Pr\!_{\widetilde{G}}(e)\ \right)
$$
}

Similarly, possible instantiations of an uncertain graph dataset $\widetilde{\mathcal{D}}=\{ \widetilde{G}_1, \cdots, \widetilde{G}_{n}\}$ are referred to as \emph{worlds} of $\widetilde{\mathcal{D}}$, where each world corresponds to an implied certain graph dataset $\mathcal{D}=\{G_1, \cdots, G_n\}$. A certain graph dataset $\mathcal{D}$ is called as being \emph{implied} from uncertain graph dataset $\widetilde{\mathcal{D}}$ (denoted as $\widetilde{\mathcal{D}}\Rightarrow \mathcal{D}$), iff  $|\mathcal{D}|=|\widetilde{\mathcal{D}}|$ and $\forall i\in\{1,\cdots, |\mathcal{D}|$\}, $\widetilde{G}_i \Rightarrow G_i$.
There are $\prod_{i=1}^{| \widetilde{\mathcal{D}}|} 2^{|E(\widetilde{G}_i)|}$ possible worlds for uncertain graph dataset $\widetilde{\mathcal{D}}$, denoted as $\mathcal{W}(\widetilde{\mathcal{D}}) = \{ \mathcal{D} \ |\  \widetilde{\mathcal{D}} \Rightarrow \mathcal{D}\}$. An uncertain graph dataset $\widetilde{\mathcal{D}}$ corresponds to a probability distribution over $\mathcal{W}(\widetilde{\mathcal{D}})$. We denote the probability of each certain graph dataset $\mathcal{D}\in \mathcal{W}(\widetilde{\mathcal{D}})$ being implied by $\widetilde{\mathcal{D}}$ as $\Pr(\widetilde{\mathcal{D}}\Rightarrow \mathcal{D})$.  By assuming that different uncertain graphs are independent from each other, we have 
{
$$ \Pr\!\left[\widetilde{\mathcal{D}} \Rightarrow \mathcal{D}\right] = \prod_{i=1}^{| \widetilde{\mathcal{D}}|} \Pr[\widetilde{G}_i \Rightarrow G_i] 
$$}

The concept of \textit{subgraph} is defined based upon certain graphs. Different from conventional subgraph mining problems where each subgraph feature can have multiple embeddings within one graph object, in our data model, each subgraph feature $g$ can only have one unique embedding within a certain graph $G$.
\begin{defn}[Subgraph]
Let $g = (V', E')$ and $G =(V, E)$ be two certain graphs. $g$ is a subgraph of $G$ (denoted as $g \subseteq G$) iff $V' \subseteq V$ and $E' \subseteq E$. We use $g\subseteq G$ to denote that graph $g$ is a subgraph of $G$. We also say that $G$ contains subgraph $g$.
\end{defn}
For an uncertain graph $\widetilde{G}$, the probability of $\widetilde{G}$ containing a subgraph feature $g$ is defined as follows:
{
\begin{align*}
\Pr(g\subseteq \widetilde{G})  &= \sum_{G\in\mathcal{W}(\widetilde{G})} \Pr(\widetilde{G}\Rightarrow G) \cdot I(g\subseteq G) \\
&= \begin{cases}
 \prod_{e\in E(g)} p(e) & \text{if $E(g) \subseteq E(\widetilde{G})$}\\
 0 & \text{otherwise}\\
 \end{cases}
\end{align*}
} 
\noindent which corresponds to the probability that a certain graph $G$ implied by $\widetilde{G}$ contains subgraph $g$.

We focus on mining a set of discriminative subgraph features to define the feature space of graph classification. It is assumed that a graph object $\widetilde{G}_i$ is
represented as a feature vector $\mathbf{x}_i=[x^1_i, \cdots, x^m_i]^{\top}$ associated with a set of subgraph features $\{g_1, \cdots, g_m\}$. Here, $x^k_i = \Pr(g_k\subseteq \widetilde{G}_i)  $ is the probability that $\widetilde{G}_i$ contains the subgraph feature $g_k$. Now suppose the full set of subgraph features in the graph dataset $\widetilde{\mathcal{D}}$ is $\mathcal{S}=\{g_1, \cdots, g_m\}$, which we use to predict the class labels of the graph objects. The full feature set $\mathcal{S}$ is very large. Only a subset of the subgraph features ($\mathcal{T} \subseteq \mathcal{S}$) is relevant to the graph classification task, which is the target feature set we want to find within uncertain graphs.  

The key issues of discriminative subgraph mining for uncertain graphs can be described as follows: 

\noindent\ \textbf{(P1)} How can one properly evaluate the discrimination scores of a subgraph feature considering the uncertainty of the graph structures?

\noindent\ \textbf{(P2)} How can one efficiently compute the probability distribution of a subgraph's discrimination score by avoiding the exhaustive enumeration of all possible worlds of the uncertain graph dataset? Moreover, since the subgraph enumeration is NP-hard, it is also infeasible to fully enumerate all the subgraph features for an uncertain graph dataset.

In the following sections, we will introduce the proposed framework for mining discriminative subgraphs from uncertain graphs.

\section{The Proposed Framework}\label{subsec:guc}
\subsection{Discrimination Score Distribution}\label{subsec:score_dist}
In this subsection, we address the problem (P1) discussed in the previous section. In conventional discriminative subgraph mining, the discrimination scores of subgraph features are usually defined for certain graph datasets, e.g., information gain and G-test score \cite{YCHY08}.  The score of a subgraph feature is  a fixed value indicating the discriminative power of the subgraph feature for the graph classification task. However, such concepts don't make sense to uncertain graph datasets, since an uncertain graph only contains a subgraph feature in a probabilistic sense.  Now we extend the concept of discriminative subgraph features in uncertain graph datasets.  Suppose we have an objective function $F(g, \mathcal{D})$ which measures the discrimination score of a subgraph $g$ in a certain graph dataset $\mathcal{D}$. The corresponding objective function on an uncertain graph dataset $\widetilde{\mathcal{D}}$  can be written as $F(g,\widetilde{\mathcal{D}})$ accordingly.  Note that $F(g,\widetilde{\mathcal{D}})$ is no longer a deterministic function. $F(g,\widetilde{\mathcal{D}})$ corresponds to a random variable over all possible outcomes of $F(g, \mathcal{D})$ ({\ie}, $\text{Range}(F)$) with probability distribution:

\begin{tabular}{ccc}
\toprule
$s_1$& $s_2$& $\cdots$ \\
\midrule
{ $\Pr[F(g,\widetilde{\mathcal{D}})=s_1]$}& {$\Pr[F(g,\widetilde{\mathcal{D}})=s_2]$}& $\cdots$\\
\bottomrule
\end{tabular}\\

 
\noindent where $s_i\in \text{Range}(F)$. 

The probability distribution of the discrimination score values can be defined as follows: 
{
$$\Pr\!\left[ F(g, \widetilde{\mathcal{D}}) = s\right] = \sum_{\mathcal{D} \in \mathcal{W}(\widetilde{\mathcal{D}})} \Pr[\widetilde{\mathcal{D}}\Rightarrow D] \cdot  I\left( F(g, \mathcal{D})=s\right)$$
}where $I(\pi)\in \{0,1\}$ is an indicator function, and $I(\pi)=1$ iff $\pi$ holds. In other words,  $\forall s\in Range(F)$, $\Pr[ F(g, \widetilde{\mathcal{D}}) = s]$ is the summation over the probabilities of all worlds of $\widetilde{\mathcal{D}}$ in which the discrimination score $F(g,\mathcal{D})$ is exactly $s$. Based on the discrimination score function on uncertain graphs, we define four statistical measures that evaluate the properties of the distribution of $F(g,\widetilde{\mathcal{D}})$ from different perspectives.

\begin{defn}[Mean-Score]
Given an uncertain graph dataset $\widetilde{\mathcal{D}}$, a subgraph feature $g$ and a discrimination score function $F(\cdot,\cdot)$, we define the expected discrimination score $\textsc{Exp}(F(g, \widetilde{\mathcal{D}}))$ as the mean score among all possible worlds of  $\widetilde{\mathcal{D}}$:
{
\begin{align*}
 \textsc{Exp}\!\left(F(g, \widetilde{\mathcal{D}})\right) &=\!\!\!\! \sum_{\mathcal{D} \in \mathcal{W}(\widetilde{\mathcal{D}})}\!\!\!\!\!\!\Pr[\widetilde{\mathcal{D}}\Rightarrow D] \cdot F(g, \mathcal{D})\\
 &= \sum_{s=-\infty}^{+\infty}\!\!\!\! s \cdot \Pr[F(g, \widetilde{\mathcal{D}}) = s]  
\end{align*}
}
\end{defn}

The mean discrimination score is the expectation of the random variable $F(g,\widetilde{\mathcal{D}})$. The expectation is usually used in conventional frequent pattern mining on uncertain datasets \cite{ZLGZ09,ZGL10}. However, it's worth noting that the expectation of discrimination scores may not be robust to extreme values. In discriminative subgraph mining, the value of a score function ({\eg}, frequency ratio\cite{JW11}, G-test score\cite{YCHY08}) can be $+\infty$. Such cases can easily dominate the computation of expectation, even if the probabilities are extremely small.  For example, suppose  we have a subgraph feature with the (score, probability) pairs as $(0.01, 99.99\%)$ and $(+\infty, 0.01\%)$. The expected score will be $+\infty$.  In order to address this problem, we either need to  bound the maximum value of the objective function like $\min(F(g,\widetilde{\mathcal{D}}), \frac{1}{\epsilon})$, or we need to introduce other statistical measures which are robust to extreme values.
\begin{defn}[Median-Score]
Given an uncertain graph dataset $\widetilde{\mathcal{D}}$, a subgraph feature $g$ and a discrimination score function $F(\cdot,\cdot)$ on certain graphs, we define the median discrimination score $\textsc{Median}(F(g, \widetilde{\mathcal{D}}))$ as the median score among all possible worlds of  $\widetilde{\mathcal{D}}$:
{\tiny
$$ \textsc{Median}\!\left(F(g, \widetilde{\mathcal{D}})\right) = \argmax_S \left\{\  \sum_{s=-\infty}^S\!\Pr\left[ F(g, \widetilde{\mathcal{D}}) = s\right] \, \, \le \frac{1}{2}\ \right\}$$
}
\end{defn}
The median score is relatively more robust to extreme values than expectation, although in  some cases the median score can still be infinite.  The same results can also hold for any quantile or $k$-th order statistic.

Another commonly used statistic is the mode score, {\ie}, the score value that has the largest probability. The mode score of a distribution means that the score is most likely to be observed within all possible worlds of $\widetilde{\mathcal{D}}$.
\begin{defn}[Mode-Score]
Given an uncertain graph dataset $\widetilde{\mathcal{D}}$, a subgraph feature $g$ and a discrimination score function $F(\cdot,\cdot)$, we define the mode discrimination score $\textsc{Mode}(F(g, \widetilde{\mathcal{D}}))$ as the score that is most likely among all possible worlds of  $\widetilde{\mathcal{D}}$:
{
$$ \textsc{Mode}\!\left(F(g, \widetilde{\mathcal{D}})\right) = \argmax_s \, \Pr\!\left[ F(g, \widetilde{\mathcal{D}}) = s\right]$$
}
\end{defn}

Next we consider the probability of a subgraph feature being observed as a discriminative pattern within all possible worlds of $\widetilde{\mathcal{D}}$, {\ie}, $ \Pr[F(g, \widetilde{\mathcal{D}})\ge \varphi]$. It is called $\varphi$-probability. The higher the value, the more likely that the subgraph feature is a discriminative pattern with a score larger or equals to a threshold  $\varphi$.
\begin{defn}[$\varphi$-Probability]
Given an uncertain graph dataset $\widetilde{\mathcal{D}}$, a subgraph feature $g$ and a discrimination score function $F(\cdot,\cdot)$, we define the $\varphi$-probability for discrimination score function $F(g, \widetilde{\mathcal{D}})$ as the sum of probabilities for all possible worlds of  $\widetilde{\mathcal{D}}$, where the score is greater than or equals to $\varphi$:
{
\begin{align*}
 \varphi\text{-}\!\Pr\!\left(F(g, \widetilde{\mathcal{D}})\right) & =  \sum_{\mathcal{D} \in \mathcal{W}(\widetilde{\mathcal{D}})} \Pr[\widetilde{\mathcal{D}}\Rightarrow D] \cdot I\left(F(g, \mathcal{D})\ge \varphi\right) \\
 & =  \sum_{s=\varphi}^{+\infty} \Pr[F(g, \widetilde{\mathcal{D}}) = s]
\end{align*}
}
\end{defn}
\normalsize

The $\varphi$-probability is robust to extreme values of the objective function. For the previous example,  we have a subgraph feature with score distribution: $(0.01, 99.99\%), (+\infty, 0.01\%)$. The $\varphi$-probability is $0.01\%$, when $\varphi = 1$.

We have already introduced four statistical measures of the distribution of a discrimination score function. Now the central problem for calculating all these measures is how to calculate $\Pr[F(g, \widetilde{\mathcal{D}}) = s]$ efficiently, which we will discuss in the following section.

\begin{table}[t]
    \centering
    {\scriptsize
    \caption{Summary of Discrimination Score Functions.}\label{tab:obj_func}
\begin{tabular}{ll}
\toprule
    Name
    & $f(n^g_+, n^g_-, n_+, n_-)$\\
\midrule
confidence   &  $\frac{n^g_+}{n^g_+ + n^g_-}$\\
frequency ratio &  $\left| \log\frac{n^g_+\cdot n_-}{n^g_-\cdot n_+} \right|$\\
G-test  & $2 n^g_+ \cdot \ln\!\frac{n^g_+\cdot n_-}{n^g_-\cdot n_+} + 2(n_+ - n^g_+) \cdot \ln\!\frac{n_-\cdot (n_+\ -\ n^g_+)}{n_+ \cdot (n_-\ -\ n^g_-)}$ \\
HSIC(linear)  & $\frac{(n^g_+\cdot n_- - n^g_-\cdot n_+)^2}{(n_+ + n_- -1)^2 (n_+ + n_-)^2} $\\
\bottomrule
\end{tabular}
}
\end{table}

\subsection{Efficient Computation}\label{subsec:dist_comp}
\normalsize
In this subsection, we address the problem (P2) discussed in Section~\ref{sec:prob_form}. Given a certain graph dataset $\mathcal{D}$, we denote the subsets of all positive graphs and all negative graphs as $\mathcal{D}_+$ and $\mathcal{D}_-$, respectively. Suppose the supports of subgraph feature $g$ in $\mathcal{D}_+$ and  $\mathcal{D}_-$ are  $n_{+}^g $ and $n_{-}^g$. $ n_{+}^g =|\{G; G\in \mathcal{D}_+, g\subseteq G\}|$.  Most of the existing discrimination score functions can be written as a function of  $ n^g_{+}$, $n^g_{-}$, $n_+$ and $n_-$:
\begin{equation}\label{eq:obj_func}
F(g,\mathcal{D}) = f\!\left(n_+^g,n_-^g, n_+, n_-\right)
\end{equation}

The definition in Eq.~\ref{eq:obj_func} covers many discrimination score functions including confidence\cite{GW10}, frequency ratio\cite{JW11}, information gain, G-test score\cite{YCHY08} and  HSIC\cite{KFY11}, as shown in Table~\ref{tab:obj_func}. For example,  frequency ratio can be written as $\text{r}(g)=|\log \frac{n^g_{+}\cdot n_- }{n^g_{-} \cdot n_+}|$. The G-test score can be written as $\text{G-test}(g) = 2 n^g_+ \cdot \ln \frac{ n^g_{+}\cdot n_-}{n^g_{-} \cdot n_+} + 2\left(n_+-n^g_{+}\right) \cdot \ln \frac{n_- \cdot (n_+- n^g_{+})}{  n_+ \cdot (n_- -n^g_{-})}$.  Because $n_+$ and $n_-$ are fixed numbers for different subgraph features, we simply use $f(n_+^g,n_-^g)$ for $f\!\left(n_+^g,n_-^g, n_+, n_-\right)$.

Based on the above definitions, we find that the number of possible outcomes of $F(g, \widetilde{\mathcal{D}})$ is bounded by $ n_+ \times n_-$, because $0\le n_+^g \le n_+$ and $0\le n_-^g\le n_-$. Thus, the probabilities $\Pr[F(g, \widetilde{\mathcal{D}}) = s]$ can be exactly computed via dynamic programming in $O(n^2)$ time, without enumerating all possible worlds of $\widetilde{\mathcal{D}}$. Instead, we can just enumerate all possible combinations of ($n_+^g, n_-^g$) and calculate the probability for each pair ($n^g_+, n^g_-$), denoted as $\Pr[ n^g_+, n^g_- , \widetilde{\mathcal{D}}] = \Pr\!\left[F(g,\widetilde{\mathcal{D}})=f(n^g_+, n^g_-)\right]$.  Then the values of $F(g, \widetilde{\mathcal{D}})$ in all possible worlds  with non-zero probabilities can be covered by the $ n_+ \times n_-$ cases.

Moreover, because different uncertain graphs are independent from each other,  we have
\begin{equation}\label{eq:pr_decomp}
\Pr[ n^g_+, n^g_- , \widetilde{\mathcal{D}}] = \Pr[ n^g_+ , \widetilde{\mathcal{D}}_+] \cdot \Pr[ n^g_-, \widetilde{\mathcal{D}}_- ] 
\end{equation}
where $\Pr[ n^g_+ , \widetilde{\mathcal{D}}_+]$ denotes the probability of the cases when there are $n_+^g$ graphs in $\widetilde{\mathcal{D}}_+$ that contain the subgraph $g$. $\Pr[ n^g_- , \widetilde{\mathcal{D}}_-]$ corresponds to the cases when there are $n_-^g$ graphs in $\widetilde{\mathcal{D}}_-$ that contain subgraph $g$.  Now we just need to compute the probabilities $\Pr[ n^g_+ , \widetilde{\mathcal{D}}_+]$ $(\forall n^g_+, 0\le n^g_+ \le n_+)$ and $\Pr[ n^g_- , \widetilde{\mathcal{D}}_-]$ $(\forall n^g_-, 0\le n^g_- \le n_-)$ separately.

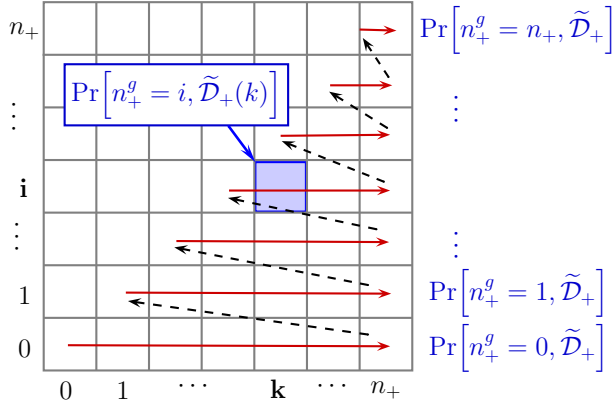
\begin{figure}[t]
\centering
\scalebox{0.7} 
{
\begin{pspicture}(2,-3.82)(9.259766,3.78)
\definecolor{color784}{rgb}{0.0,0.0,0.8}
\definecolor{color344}{rgb}{0.8,0.0,0.0}
\definecolor{color345}{rgb}{0.5,0.5,0.5}
\definecolor{color915b}{rgb}{0.8,0.8,1.0}
\psframe[linewidth=0.04,linecolor=blue,dimen=outer,fillstyle=solid,fillcolor=color915b](5.478906,0.74)(4.47890624,-0.24)
\psline[linewidth=0.04cm,linecolor=color345](1.4789063,-3.24)(1.4789063,3.76)
\psline[linewidth=0.04cm,linecolor=color345](2.4789062,-3.24)(2.4789062,3.76)
\psline[linewidth=0.04cm,linecolor=color345](3.4789062,-3.24)(3.4789062,3.76)
\psline[linewidth=0.04cm,linecolor=color345](4.478906,-3.24)(4.478906,3.76)
\psline[linewidth=0.04cm,linecolor=color345](5.478906,-3.24)(5.478906,3.76)
\psline[linewidth=0.04cm,linecolor=color345](7.478906,1.76)(0.47890624,1.76)
\psline[linewidth=0.04cm,linecolor=color345](7.478906,0.76)(0.47890624,0.76)
\psline[linewidth=0.04cm,linecolor=color345](7.478906,-0.24)(0.47890624,-0.24)
\psline[linewidth=0.04cm,linecolor=color345](7.478906,-1.24)(0.47890624,-1.24)
\psline[linewidth=0.04cm,linecolor=color345](7.478906,-2.24)(0.47890624,-2.24)
\usefont{T1}{ptm}{b}{n}\rput(0.9014453,-3.615){\Large $0$}
\usefont{T1}{ptm}{b}{n}\rput(1.9846874,-3.615){\Large $1$}
\usefont{T1}{ptm}{b}{n}\rput(6.9771485,-3.615){\Large $n_+$}
\psline[linewidth=0.04cm,linecolor=color345](7.458906,2.76)(0.45890626,2.76)
\psline[linewidth=0.04cm,linecolor=color345](6.478906,-3.24)(6.478906,3.76)
\usefont{T1}{ptm}{b}{n}\rput(0.11714844,-2.815){\Large $0$}
\usefont{T1}{ptm}{b}{n}\rput(0.11714844,-1.815){\Large $1$}
\usefont{T1}{ptm}{b}{n}\rput(0.11714844,3.205){\Large $n_+$}
\psframe[linewidth=0.04,linecolor=color345,dimen=outer](7.498906,3.78)(0.45890626,-3.26)
\usefont{T1}{ptm}{b}{n}\rput(4.9266405,-3.595){\Large $\mathbf{k}$}
\usefont{T1}{ptm}{b}{n}\rput(0.09808594,0.225){\Large $\mathbf{i}$}
\psline[linewidth=0.04cm,linecolor=color344,arrowsize=0.1cm 2.0,arrowlength=1.4,arrowinset=0.4]{->}(0.93890625,-2.76)(7.058906,-2.78)
\psline[linewidth=0.04cm,linecolor=color344,arrowsize=0.1cm 2.0,arrowlength=1.4,arrowinset=0.4]{->}(2.0389063,-1.76)(7.078906,-1.78)
\psline[linewidth=0.04cm,linecolor=color344,arrowsize=0.1cm 2.0,arrowlength=1.4,arrowinset=0.4]{->}(2.9989061,-0.78)(7.038906,-0.8)
\psline[linewidth=0.04cm,linecolor=color344,arrowsize=0.1cm 2.0,arrowlength=1.4,arrowinset=0.4]{->}(3.9989061,0.18)(7.058906,0.18)
\psline[linewidth=0.04cm,linecolor=color344,arrowsize=0.1cm 2.0,arrowlength=1.4,arrowinset=0.4]{->}(4.978906,1.22)(7.058906,1.24)
\psline[linewidth=0.04cm,linecolor=color344,arrowsize=0.1cm 2.0,arrowlength=1.4,arrowinset=0.4]{->}(5.918906,2.18)(7.098906,2.18)
\psline[linewidth=0.04cm,linecolor=color344,arrowsize=0.1cm 2.0,arrowlength=1.4,arrowinset=0.4]{->}(6.478906,3.24)(7.1789064,3.22)
\psline[linewidth=0.04cm,linestyle=dashed,dash=0.16cm 0.16cm,arrowsize=0.1cm 2.0,arrowlength=1.4,arrowinset=0.4]{cc->}(6.6589065,-2.56)(2.0789063,-1.92)
\psline[linewidth=0.04cm,linestyle=dashed,dash=0.16cm 0.16cm,arrowsize=0.1cm 2.0,arrowlength=1.4,arrowinset=0.4]{cc->}(6.6589065,-1.62)(2.9589062,-0.9)
\psline[linewidth=0.04cm,linestyle=dashed,dash=0.16cm 0.16cm,arrowsize=0.1cm 2.0,arrowlength=1.4,arrowinset=0.4]{cc->}(6.8189063,-0.56)(3.9789062,0.04)
\psline[linewidth=0.04cm,linestyle=dashed,dash=0.16cm 0.16cm,arrowsize=0.1cm 2.0,arrowlength=1.4,arrowinset=0.4]{cc->}(6.958906,0.34)(4.998906,1.12)
\psline[linewidth=0.04cm,linestyle=dashed,dash=0.16cm 0.16cm,arrowsize=0.1cm 2.0,arrowlength=1.4,arrowinset=0.4]{cc->}(7.018906,1.36)(5.898906,2.06)
\psline[linewidth=0.04cm,linestyle=dashed,dash=0.16cm 0.16cm,arrowsize=0.1cm 2.0,arrowlength=1.4,arrowinset=0.4]{cc->}(7.038906,2.32)(6.538906,3.08)
\usefont{T1}{ptm}{b}{n}\rput(5.9979296,-3.535){\Large $\cdots$}
\usefont{T1}{ptm}{b}{n}\rput(3.3579297,-3.535){\Large $\cdots$}
\usefont{T1}{ptm}{b}{n}\rput{-270.0}(1.6921874,1.6141406){\rput(0.017929688,1.765){\Large $\cdots$}}
\usefont{T1}{ptm}{b}{n}\rput{-270.0}(-0.5278125,-0.7258594){\rput(0.07792969,-0.515){\Large $\cdots$}}
\usefont{T1}{ptm}{b}{n}\rput(9.5,-2.815){\Large \color{color784}$\Pr\!\left[n_+^g = 0,\widetilde{\mathcal{D}}_+\right]$}
\usefont{T1}{ptm}{b}{n}\rput{-270.0}(7.6521873,-9.225859){\rput(8.41793,-0.675){\Large \color{color784}$\cdots$}}
\usefont{T1}{ptm}{b}{n}\rput{-270.0}(7.6521873,-9.225859){\rput(11,-0.675){\Large \color{color784}$\cdots$}}
\usefont{T1}{ptm}{b}{n}\rput(9.5,-1.775){\Large \color{color784}$\Pr\!\left[n_+^g = 1,\widetilde{\mathcal{D}}_+\right]$}
\usefont{T1}{ptm}{b}{n}\rput(9.5,3.225){\Large \color{color784}$\Pr\!\left[n_+^g = n_+,\widetilde{\mathcal{D}}_+\right]$}
\rput(3,2){\Large \color{color784}\psframebox[linewidth=0.04,linecolor=color784,fillstyle=solid]{$\Pr\!\left[n_+^g =i,\widetilde{\mathcal{D}}_+(k)\right]$}}
\psline[linewidth=0.05cm,linecolor=blue,arrowsize=0.08cm 2.0,arrowlength=2,arrowinset=0.4]{cc->}(3.9789062,1.42)(4.498906,0.72)
\end{pspicture} 
}
\caption{The dynamic programming process for computing $\Pr\!\left[n_+^g, \widetilde{\mathcal{D}}_+\right]$. The same process applies for $\Pr\!\left[n_-^g, \widetilde{\mathcal{D}}_-\right]$.} \label{fig:dp}
\end{figure}


Let  $\widetilde{\mathcal{D}}(k)$ denote the first $k$ uncertain graphs in $\widetilde{\mathcal{D}}$, {\ie}, $\widetilde{\mathcal{D}}(k)= \{ \widetilde{G}_1, \cdots, \widetilde{G}_k \}$. $\widetilde{\mathcal{D}}_+(k)$ and $\widetilde{\mathcal{D}}_-(k)$ denote the first $k$ graphs in $\widetilde{\mathcal{D}}_+$ and $\widetilde{\mathcal{D}}_-$ respectively. All the values of $\Pr[ n^g_+ , \widetilde{\mathcal{D}}_+]$ and  $\Pr[ n^g_- , \widetilde{\mathcal{D}}_-]$ can be calculated using the recursive equation in Figure~\ref{fig:recursive}. The  $\Pr[ i ,\widetilde{\mathcal{D}}({k})]$ denotes the probability when there are $i$ graphs containing $g$ in $\widetilde{\mathcal{D}}({k})$. And the target values to calculate are $\Pr[ i ,\widetilde{\mathcal{D}}_{+}(n_+) ]$ ($\forall i, 0\le i \le n_+$) and $\Pr[ i ,\widetilde{\mathcal{D}}_{-}(n_-) ]$ ($\forall i, 0\le i \le n_-$) by substituting the $\widetilde{\mathcal{D}}_+$ and $\widetilde{\mathcal{D}}_-$ into the Eq.~\ref{eq_dp}, respectively. In Figure~\ref{fig:algorithm_dp}, we showed the dynamic programing algorithm to compute the target values using Eq.~\ref{eq_dp}. Figure~\ref{fig:dp} illustrates the computation process of the dynamic programing algorithm for $\Pr[ n^g_+,\widetilde{\mathcal{D}}_{+}] $, while the same process also applies for $\Pr[ n^g_-,\widetilde{\mathcal{D}}_{-}] $.

For  details of the recursive equations in Figure~\ref{fig:recursive}, we have the base cases, $\Pr[0 , \widetilde{\mathcal{D}}_0] =1$ and  $\Pr[ i , \widetilde{\mathcal{D}}(k)]=0$ (if $i>k$ or $i<0$). For other cases, the probability value can be calculated through the recursive equation in Eq.~\ref{eq_dp}. Then,  $\Pr[ n^g_+, n^g_- , \widetilde{\mathcal{D}}]$ can be calculated via Eq.~\ref{eq:pr_decomp}. Thus all the statistical measures mentioned in Section~\ref{subsec:score_dist} can be calculated within $O(n^2)$ time as follows:
$$ \textsc{Exp}\!\left(F(g, \widetilde{\mathcal{D}})\right)  = \sum_{n_+^g=0}^{n_+}\sum_{n_-^g=0}^{n_-}  \Pr[n_+^g, n_-^g, \widetilde{\mathcal{D}})] \cdot f(n_+^g, n_-^g)$$
{\scriptsize
$$ \textsc{Median}\!\left(F(g, \widetilde{\mathcal{D}})\right) = \argmax_s\! \left\{ \sum_{x=-\infty}^s\sum_{f(n_+^g, n_-^g)=x}\!\!\!\!\!\!Pr[n_+^g, n_-^g, \widetilde{\mathcal{D}})]  \ \, \le\! \frac{1}{2}\right\}$$
$$  \textsc{Mode}\!\left(F(g, \widetilde{\mathcal{D}})\right)\! =\! \argmax_s\!\sum_{n_+^g=0}^{n_+}\sum_{n_-^g=0}^{n_-} \Pr[n_+^g, n_-^g, \widetilde{\mathcal{D}})] \cdot I(f(n_+^g, n_-^g)\!=\!s)$$
}
{\small
$$  \varphi\text{-}\!\Pr\!\left(F(g, \widetilde{\mathcal{D}})\right)\! =\! \sum_{n_+^g=0}^{n_+}\sum_{n_-^g=0}^{n_-} \Pr[n_+^g, n_-^g, \widetilde{\mathcal{D}})] \cdot I( f(n_+^g, n_-^g)\!\ge\! \varphi)$$
}

We will show later  that the dynamic programming process is highly efficient in all the applications studied in Section~\ref{sec:experiment}. For dataset with even larger number of graphs, the divid-and-conquer method in \cite{SCCC10} could also be used here to further optimize the computational cost. 

\begin{figure}[t]
\center
{\scriptsize
\begin{tabular}{l}
\toprule
  \textbf{Input:}\\
  \begin{tabular}{r@{: }l}
    $\widetilde{\mathcal{D}}_{+}$ & the set of positive graphs\\
    $\widetilde{\mathcal{D}}_{-}$ & the set of negative graphs\\
  \end{tabular}\\
  \midrule
\textbf{Dynamic Programming:}\\
\quad for $n_+^g \leftarrow 0$ to $n_+$\\
\quad \quad for $k\leftarrow n_+^g $ to $n_+$\\
\quad \qquad compute $\Pr[\ n_+^g, \widetilde{\mathcal{D}}_{+}(k)\ ]$ via Eq.~\ref{eq_dp};\\
\quad for $n_-^g \leftarrow 0$ to $n_-$\\
\quad \quad for $k\leftarrow n_-^g $ to $n_-$\\
\quad \qquad compute $\Pr[\ n_-^g ,  \widetilde{\mathcal{D}}_{-}(k)\ ]$ via Eq.~\ref{eq_dp};\\
  \textbf{Output:}\\
  \begin{tabular}{rl}
      $\Pr[n_+^g ,  \widetilde{\mathcal{D}}_{+}\ ]$ ($\forall \ n_+^g,\ 0\le n_+^g \le n_+$)\\
      $\Pr[n_-^g ,  \widetilde{\mathcal{D}}_{-}\ ]$ ($\forall \ n_-^g,\ 0\le n_-^g \le n_-$)\\
    \end{tabular}\\
    \bottomrule
\end{tabular}
}
\caption{The dynamic programming algorithm for probability computation.}\label{fig:algorithm_dp}
\end{figure}

\begin{figure*}[t]
{\scriptsize
\begin{equation}
\Pr\!\left[ i, \widetilde{\mathcal{D}}(k)\right] = 
\begin{cases}
\left(1- \Pr[g\subseteq \widetilde{G}_k]\right)\cdot \Pr[ i , \widetilde{\mathcal{D}}(k-1) ]\ +\ \Pr[g\subseteq \widetilde{G}(k)] \cdot\Pr[ i-1 , \widetilde{\mathcal{D}}(k-1) ] & \text{if } i\le k\\
1 & \text{if } i=k=0\\
0 & \text{if } i>k  \text{ or } i<0
\end{cases}\label{eq_dp}
\end{equation}
}
\caption{Recursive equation for dynamic programming.}\label{fig:recursive}
\end{figure*}

\begin{figure*}[!ht]
\center
{\scriptsize
\begin{tabular}{l}
\toprule
  \textbf{Input:}\\
  \begin{tabular}{r@{: }l@{\qquad}r@{: }l}
    $\widetilde{\mathcal{D}}$ & the uncertain graph dataset $\{\widetilde{G}_1, \cdots, \widetilde{G}_n\}$ & $t$ & the maximum number of subgraphs.\\
    $\mathbf{y}$ & the vector of class labels for uncertain graphs, & $min\_sup$ & the minimum expected frequency.\\
    $M$& the statistic measure (Expectation/Median/Mode/$\varphi$-Pr)
  \end{tabular}\vspace{3pt}\\
\textbf{Recursive Subgraphs Mining:}\\
\quad - Depth-first search the gSpan's code tree and update the feature list as follows:\\
     \qquad 1. Update the candidate feature list using the current subgraph feature $g_c$:\\
     \qquad \qquad Calculate the probability vector $\Pr[n_+^{g_c} ,  \widetilde{\mathcal{D}}_{+}\ ]$ and $\Pr[n_-^{g_c} ,  \widetilde{\mathcal{D}}_{-}\ ]$ using the dynamic programing algorithm in Figure~\ref{fig:algorithm_dp}\\
     \qquad \qquad Compute the statistic measure $M\left(F(g_c, \widetilde{\mathcal{D}})\right)$ based on the discrimination score function $F(g_c, \widetilde{\mathcal{D}})$. \\
      \qquad \qquad If the score is larger than the worst feature in $\mathcal{T}$, replace it and update $\theta = \min_{g\in \mathcal{T}} M\left(F(g, \widetilde{\mathcal{D}})\right)$\\
     \qquad 2. Test pruning criteria for the sub-tree rooted from node $g$ as follows:\\
     \qquad \qquad if $\text{Exp-Freq}(g_c)\le min\_sup$, prune the sub-tree of $g_c$\\
     \qquad \qquad  if $\text{Bound-}M\left(F(g_c, \widetilde{\mathcal{D}})\right)\le \theta$, prune the sub-tree of $g_c$\\
     \qquad 3. Recursion: Depth-first search the sub-tree rooted from node $g_c$\vspace{2pt}\\
  \textbf{Output:}\\
  \begin{tabular}{rl}
      $\mathcal{T}$: & the discriminative subgraph features for uncertain graph classification.\\
    \end{tabular}\\
    \bottomrule
  \end{tabular}
  }
  \caption{The \textnormal{\textsc{Dug}} framework for discriminative subgraph mining.}\label{fig:algorithm_guc}
\end{figure*}

\subsection{Upper-Bounds for Subgraph Pruning}\label{subsec:bound}

In order to avoid the exhaustive enumeration of subgraph features, we derive some subgraph pruning methods. One natural pruning bound for subgraph search is the expected frequency of a subgraph feature, $\text{Exp-Freq}(g,\widetilde{\mathcal{D}}) = \frac{\sum_i^n \Pr (g\subseteq \widetilde{\mathcal{G}}_i)}{n}$, since it's can be easily proved with anti-monotonic property. 
For the expectation and $\varphi$-probability, we can also derive additional bounds for subgraph pruning. Let $\hat{F}(g,\mathcal{D}) = \hat{f}(n_+^g, n_-^g)$ be the estimated upper-bound function for $g$ and its supergraphs in certain graph dataset $\mathcal{D}$. We can derive the corresponding upper-bounds as follows:
$$ \textsc{UB-Exp}(g, \widetilde{\mathcal{D}})  = \sum_{n_+^g=0}^{n_+}\sum_{n_-^g=0}^{n_-}  \Pr[n_+^g, n_-^g, \widetilde{\mathcal{D}})] \cdot \hat{f}(n_+^g, n_-^g)$$
{\small
$$  \textsc{UB-}\varphi\text{-}\!\Pr(g, \widetilde{\mathcal{D}}) = \sum_{n_+^g=0}^{n_+}\sum_{n_-^g=0}^{n_-} \Pr[n_+^g, n_-^g, \widetilde{\mathcal{D}})] \cdot I( \hat{f}(n_+^g, n_-^g)\ge \varphi)$$
}
For the median and mode measures, it is difficult to derive a meaningful bound, thus we simply use the expected frequency to perform the subgraph pruning.

We now utilize the above bounds to prune the DFS-code tree in gSpan \cite{YH02} by the branch-and-bound pruning.  The top-$t$ best features are maintained in a candidate list. During the subgraph mining, we calculate the upper-bound of each subgraph feature in the search tree. If a subgraph feature with its children pattern cannot update the candidate feature list, we can prune the subtree of gSpan rooted from this node. It is guaranteed by the upper-bounds that we will not miss any better subgraph features. Thus, the subgraph mining process can be speeded up without loss of performance. The algorithm of {\guc} is summarized in Figure~\ref{fig:algorithm_guc}.


\section{Experiments}\label{sec:experiment}


In order to evaluate the performance of the proposed approach for uncertain graph classification, we tested our algorithm on real-world fMRI brain images as summarized in Table~\ref{tab:datastat}. 

\subsection{Data Collection}
In order to evaluate the performance of the proposed approach for uncertain graph classification, we tested our algorithm on real-world fMRI brain images. 

\noindent\textbullet\ \textit{Alzheimer's Disease (ADNI)}: The first dataset is collected from the Alzheimer's Disease Neuroimaging Initiative\footnote{http://adni.loni.ucla.edu/}. The dataset consists of records of patients with Alzheimer's Disease (AD) and Mild Cognitive Impairment (MCI).  We downloaded all records of resting-state fMRI images and treated the normal brains as \emph{negative} graphs, and  AD+MCI as the \emph{positive} graphs.  We applyed Automated Anatomical Labeling (AAL\footnote{\url{http://www.cyceron.fr/web/aal__anatomical_automatic_labeling.html}}) to extract a sequence of responds from each of of the $116$ anatomical volumes of interest (AVOI), where each AVOI represents a different brain region. The correlations of  brain activities  among different brain regions are computed. Positive correlations are used as uncertain links among brain regions. For details, we used SPM8 toolbox\footnote{http://www.fil.ion.ucl.ac.uk/spm/software/spm8/}, and functional images were realigned to the first volume, slice timing corrected, and normalized to the MNI template and spatially smoothed with an 8-mm Gaussian kernel. Resting-State fMRI Data Analysis Toolkit (REST\footnote{http://resting-fmri.sourceforge.net}) was then used to remove the linear trend of time series and temporally band-pass filtering (0.01-0.08 Hz). Before the correlation analysis, several sources of spurious variance were then removed from the data through linear regression: (i) six parameters obtained by rigid body correction of head motion, (ii) the whole-brain signal averaged over a fixed region in atlas space, (iii) signal from a ventricular region of interest, and (iv) signal from a region centered in the white matter.  Each brain is represented as an uncertain graph with 90 nodes corresponding to 90 cerebral regions, excluding 26 cerebellar regions. 

\noindent\textbullet\  \textit{Attention Deficit Hyperactivity Disorder (ADHD)}: The second dataset is collected from ADHD-200 global competition dataset \footnote{http://neurobureau.projects.nitrc.org/ADHD200/}. The dataset contains records of resting-state fMRI images for 776 subjects, which are labeled as real patients (positive) and normal controls (negative). Similar to the ADNI dataset, the brain images are preprocessed using Athena Pipeline\footnote{\url{http://www.nitrc.org/plugins/mwiki/index.php/neurobureau:AthenaPipeline}}. The original dataset is unbalanced, we randomly sampled 100 ADHD patients and 100 normal controls from the dataset for performance evaluation.

\noindent\textbullet\  \textit{Human Immunodeficiency Virus Infection (HIV)}: The third dataset is collected from the Chicago Early HIV Infection Study in Northwestern University \cite{WFOC11}. The dataset contains fMRI brain images of patients with early HIV infection (positive) as well as normal controls (negative). The same preprocessing steps as in ADNI dataset were used to extract a functional connectivity network from each image.

\begin{table}[!t]
    \centering
    \caption{Summary of experimental datasets. }\label{tab:datastat}
{\scriptsize
\begin{tabular}{rcccccc}
\toprule
    & $|\widetilde{\mathcal{D}}|$
    & $|\widetilde{\mathcal{D}}_+|$
    & $|\widetilde{\mathcal{D}}_-|$
    & $|V|$
    & avg. $|E|$
    & avg. edge prob \\
\midrule
ADHD&200&100&100&116&484.7&0.55\\
ADNI&36&18&18&90&2019.8&0.59\\
HIV&50&25&25&90&480.48&0.88\\
\bottomrule
\end{tabular}
}
\end{table}

\begin{table*}[t]
\caption{Results on the ADNI (Alzheimer's Disease) dataset with different number of features($t=100, \cdots, 500$). The results are reported as ``average performance + (rank)". }\label{tab:adni}
{\tiny
\centering
\begin{tabular}{rlllllrlllllr}
\toprule
 & \multicolumn{5}{c}{Error Rate $\downarrow$}&&\multicolumn{5}{c}{F1 $\uparrow$}&Avg.\\
\cmidrule{2-6}\cmidrule{8-12}
{Methods}&\multicolumn{1}{c}{$t=100$}&\multicolumn{1}{c}{$t=200$}&\multicolumn{1}{c}{$t=300$}&\multicolumn{1}{c}{$t=400$}&\multicolumn{1}{c}{$t=500$}&&\multicolumn{1}{c}{$t=100$}&\multicolumn{1}{c}{$t=200$}&\multicolumn{1}{c}{$t=300$}&\multicolumn{1}{c}{$t=400$}&\multicolumn{1}{c}{$t=500$}&\multicolumn{1}{c}{Rank}\\
\midrule
{\fexp Exp}-HSIC 				&0.400 {\frank (9)} 		&	0.367 {\frank (8)} 	&0.367 {\frank (10)}&0.317 {\frank (4)} 		&0.333 {\frank (9)} 		&&0.699 {\frank (9)} 	&0.725 {\frank (9)} 		&0.725 {\frank (9)} 	&0.753 {\frank (6)} 		&0.743 {\frank (10)} 		&{\frank (8.3)}\\
{\fmed Med}-HSIC  				&0.433 {\frank (14)} 	&	0.350 {\frank (5)} 	&0.333 {\frank (6)} 	&0.350 {\frank (8)} 		&0.317 {\frank (7)} 		&&0.667 {\frank (13)} 	&0.741 {\frank (7)} 		&0.757 {\frank (4)} 	&0.734 {\frank (9)} 		&0.766 {\frank (7)} 		&{\frank (8.0)}\\
{\fmod Mod}-HSIC 				&0.367 {\frank (6)} 		&	0.333 {\frank (3)} 	&0.300 {\frank (1)*} 	&0.317 {\frank (4)} 		&0.300 {\frank (2)} 		&&0.703 {\frank (8)} 	&0.750 {\frank (4)} 		&0.776 {\frank (3)} 	&0.766 {\frank (3)} 		&0.775 {\frank (4)} 		&{\frank (3.8)}\\
{\fphi $\varphi$Pr}-HSIC	&0.283 {\frank (1)*}	&	0.283 {\frank (1)*} 	&0.333 {\frank (6)} 	&0.333 {\frank (7)} 		&0.300 {\frank (2)} 		&&0.778 {\frank (1)*}	&0.785 {\frank (1)*}	&0.757 {\frank (4)} 	&0.750 {\frank (7)} 	&0.776 {\frank (3)} 		&{\frank (3.3)}\\
{\fbaseline HSIC}	&0.450 {\frank (16)}	&		0.467 {\frank (19)} 	&	0.467 {\frank (17)} 	&	0.500 {\frank (18)} 		&	0.500 {\frank (18)} 		&&0.615 {\frank (18)}	&	0.597 {\frank (19)}	&	0.622 {\frank (17)} 	&	0.583 {\frank (18)} 	&0.584 {\frank (18)} 		&{\frank (18.1)}\\
\cmidrule{2-6}\cmidrule{8-12}
{\fexp Exp}-Ratio 				&0.433 {\frank (14)} 	&	0.383 {\frank (10)} 	&0.317 {\frank (4)} 	&0.300 {\frank (2)} 		&0.300 {\frank (2)} 		&&0.667 {\frank (13)} 	&0.715 {\frank (10)}	&0.756 {\frank (6)} 	&0.766 {\frank (3)} 		&0.766 {\frank (7)} 		&{\frank (7.1)}\\
{\fmed Med}-Ratio 				&0.450 {\frank (16)} 	&	0.417 {\frank (15)} 	&0.450 {\frank (16)}&0.383 {\frank (11)} 	&0.383 {\frank (11)} 		&&0.639 {\frank (17)} 	&0.653 {\frank (16)}	&0.608 {\frank (20)} &0.689 {\frank (12)} 	&0.684 {\frank (11)} 		&{\frank (14.5)}\\
{\fmod Mod}-Ratio 				&0.317 {\frank (3)} 		&	0.350 {\frank (5)} 	&0.433 {\frank (15)}&0.417 {\frank (13)} 	&0.467 {\frank (15)} 		&&0.776 {\frank (2)} 	&0.744 {\frank (6)} 		&0.659 {\frank (13)}&0.657 {\frank (13)} 	&0.612 {\frank (15)} 		&{\frank (9.9)}\\
{\fphi $\varphi$Pr}-Ratio	&0.400 {\frank (9)} 		&	0.317 {\frank (2)} 	&0.300 {\frank (1)*} 	&0.300 {\frank (2)} 		&0.267 {\frank (1)*} 		&&0.692 {\frank (10)} 	&0.764 {\frank (2)} 		&0.784 {\frank (1)*} 	&0.778 {\frank (2)} 	&0.809 {\frank (1)*} 		&{\frank (3.1)}\\
{\fbaseline Ratio} 				&0.500 {\frank (19)} 	&		0.483 {\frank (20)} 	&	0.533 {\frank (22)} 	&	0.567 {\frank (22)} 		&	0.533 {\frank (20)} 		&&0.581 {\frank (20)} 	&	0.603 {\frank (18)}	&	0.533 {\frank (21)} 	&	0.519 {\frank (22)} 		&	0.550 {\frank (20)} 		&{\frank (20.4)}\\
\cmidrule{2-6}\cmidrule{8-12}
{\fexp Exp}-Gtest 				&0.300 {\frank (2)} 		&	0.367 {\frank (8)} 	&0.317 {\frank (4)} 	&0.350 {\frank (8)} 		&0.383 {\frank (11)} 		&&0.774 {\frank (3)} 	&0.693 {\frank (11)}	&0.729 {\frank (9)} 	&0.702 {\frank (10)} 	&0.672 {\frank (12)} 		&{\frank (7.8)}\\
{\fmed Med}-Gtest 				&0.517 {\frank (21)} 	&	0.450 {\frank (18)} 	&0.400 {\frank (11)}&0.500 {\frank (18)} 	&0.483 {\frank (17)} 		&&0.562 {\frank (21)} 	&0.597 {\frank (19)}	&0.655 {\frank (14)}&0.567 {\frank (19)} 	&0.589 {\frank (17)} 		&{\frank (17.5)}\\
{\fmod Mod}-Gtest 				&0.517 {\frank (21)} 	&	0.550 {\frank (22)} 	&0.500 {\frank (21)}&0.500 {\frank (18)} 	&0.517 {\frank (19)} 		&&0.531 {\frank (22)} 	&0.491 {\frank (22)}	&0.527 {\frank (22)}&0.545 {\frank (20)} 	&0.558 {\frank (19)} 		&{\frank (20.6)}\\
$\varphi$Pr-Gtest	&0.450 {\frank (16)} 	&	0.417 {\frank (15)} 	&0.417 {\frank (13)}&0.383 {\frank (11)} 	&0.300 {\frank (2)} 		&&0.648 {\frank (16)} 	&0.675 {\frank (14)}	&0.665 {\frank (12)}&0.701 {\frank (11)} &0.768 {\frank (6)} 		&{\frank (11.6)}\\
{\fbaseline Gtest} 				&0.500 {\frank (19)} 		&		0.500 {\frank (21)} 	&	0.467 {\frank (17)} 	&	0.433 {\frank (14)} 		&	0.550 {\frank (21)}  &&0.583 {\frank (19)} 	&	0.580 {\frank (21)}	&	0.612 {\frank (19)} 	&	0.656 {\frank (14)} 	&	0.547 {\frank (21)} 		&{\frank (18.6)}\\
\cmidrule{2-6}\cmidrule{8-12}
{\fexp Exp}-Conf 				&0.367 {\frank (7)} 		&	0.333 {\frank (3)} 	&0.300 {\frank (1)*} 	&0.283 {\frank (1)*} 		&0.300 {\frank (2)} 		&&0.744 {\frank (6)} 	&0.762 {\frank (3)} 		&0.780 {\frank (2)} 	&0.795 {\frank (1)*} 		&0.780 {\frank (2)} 		&{\frank *(2.8)}\\
{\fmed Med}-Conf 				&0.333 {\frank (4)} 		&	0.350 {\frank (5)} 	&0.350 {\frank (8)} 	&0.350 {\frank (8)} 		&0.317 {\frank (7)} 		&&0.760 {\frank (4)} 	&0.747 {\frank (5)} 		&0.752 {\frank (7)} 	&0.740 {\frank (8)} 		&0.770 {\frank (5)} 		&{\frank (6.1)}\\
{\fmod Mod}-Conf 				&0.417 {\frank (12)} 	&	0.383 {\frank (10)} 	&0.350 {\frank (8)} 	&0.317 {\frank (4)} 		&0.333 {\frank (9)} 		&&0.690 {\frank (11)} 	&0.728 {\frank (8)} 		&0.742 {\frank (8)} 	&0.759 {\frank (5)} 		&0.750 {\frank (9)} 		&{\frank (8.4)}\\
{\fphi $\varphi$Pr}-Conf   &0.400 {\frank (9)} 		&	0.417 {\frank (15)} 	&0.467 {\frank (17)}&0.467 {\frank (16)} 	&0.433 {\frank (13)} 		&&0.685 {\frank (12)} 	&0.648 {\frank (17)}	&0.619 {\frank (18)}&0.592 {\frank (17)} &0.632 {\frank (13)} 		&{\frank (14.7)}\\
{\fbaseline Conf}	&0.400 {\frank (9)} 		&		0.400 {\frank (13)} 	&	0.417 {\frank (13)} 	&	0.450 {\frank (15)} 		&	0.467 {\frank (15)} 		&&0.655 {\frank (15)} 	&	0.667 {\frank (15)} 		&	0.645 {\frank (15)} 	&	0.618 {\frank (15)} 		&	0.610 {\frank (16)} 		&{\frank (14.1)}\\
\cmidrule{2-6}\cmidrule{8-12}
Exp-Freq	&0.383 {\frank (8)} 		&	0.383 {\frank (10)} 	&0.400 {\frank (11)}&0.467 {\frank (16)} 	&0.433 {\frank (13)} 		&&0.705 {\frank (7)} 	&0.685 {\frank (13)}&0.675 {\frank (11)} 	&0.607 {\frank (16)} 	&0.632 {\frank (13)} 		&{\frank (11.8)}\\
{\fbaseline Freq }				&0.350 {\frank (5)} 		&		0.400 {\frank (13)} 	&	0.483 {\frank (20)}&	0.550 {\frank (21)} 	&	0.550 {\frank (21)} 		&&0.747 {\frank (5)} 	&	0.692 {\frank (12)}&	0.627 {\frank (16)} 	&	0.539 {\frank (21)} 	&	0.547 {\frank (21)} 		&{\frank (15.5)}\\
\bottomrule
\end{tabular}
}
\end{table*}

\begin{table*}[t]
\caption{Results on the ADHD (Attention Deficit Hyperactivity Disorder) dataset with different number of features ($t=100, \cdots, 500$). The results are reported as ``average performance + (rank)". }\label{tab:adhd}
{\tiny
\centering
\begin{tabular}{rlllllrlllllr}
\toprule
 & \multicolumn{5}{c}{Error Rate $\downarrow$}&&\multicolumn{5}{c}{F1 $\uparrow$}&Avg.\\
\cmidrule{2-6}\cmidrule{8-12}
{Methods}&\multicolumn{1}{c}{$t=100$}&\multicolumn{1}{c}{$t=200$}&\multicolumn{1}{c}{$t=300$}&\multicolumn{1}{c}{$t=400$}&\multicolumn{1}{c}{$t=500$}&&\multicolumn{1}{c}{$t=100$}&\multicolumn{1}{c}{$t=200$}&\multicolumn{1}{c}{$t=300$}&\multicolumn{1}{c}{$t=400$}&\multicolumn{1}{c}{$t=500$}&\multicolumn{1}{c}{Rank}\\
\midrule
{\fexp Exp}-HSIC 				&0.423 {\frank (10)}	&0.438 {\frank (13)} 	&0.455 {\frank (14)} 	&0.455 {\frank (11)} 	&	0.448{\frank (12)} 		&&0.593 {\frank (10)} &	0.564 {\frank (13)}	&0.543 {\frank (14)} 	&0.547 {\frank (11)}	&	0.549 {\frank (12)}	&{\frank (12.0)}\\
{\fmed Med}-HSIC  				&0.420 {\frank (9)} 		&0.405 {\frank (8)} 		&0.413 {\frank (8)} 		&0.448 {\frank (10)} 	&	0.433 {\frank (6)} 		&&0.569 {\frank (13)} &	0.597 {\frank (7)} 	&0.593 {\frank (5)} 		&0.549 {\frank (10)}	&	0.562 {\frank (7)}		&{\frank (8.3)}\\
{\fmod Mod}-HSIC 				&0.390 {\frank (4)} 		&0.405 {\frank (8)} 		&	0.403 {\frank (4)} 	&0.393 {\frank (1)*}	&0.410 {\frank (2)} 			&&0.614 {\frank (3)} 	&	0.599 {\frank (6)} 	&0.596 {\frank (4)} 		&0.594 {\frank (1)*}	&0.584 {\frank (2)}		&{\frank *(3.5)}\\
{\fphi $\varphi$Pr}-HSIC	&0.432 {\frank (12)} 	&0.470 {\frank (17)} 	&0.475 {\frank (16)} 	&0.513 {\frank (22)} 	&	0.503 {\frank (21)} 		&&0.597 {\frank (7)} 	&	0.563 {\frank (14)}	&0.554 {\frank (13)} 	&0.508 {\frank (17)}	&	0.525 {\frank (18)}	&{\frank (15.7)}\\
{\fbaseline HSIC}	&0.529 {\frank (22)}	&	0.510 {\frank (20)} 	&		0.488 {\frank (17)} 	&	0.455 {\frank (11)} 		&	0.485 {\frank (17)} 		&&	0.505 {\frank (22)}	&	0.494{\frank (18)}	& 0.498 {\frank (18)} 	&	0.538   {\frank (13)} 	&0.526 {\frank (17)} 		&{\frank (17.5)}\\
\cmidrule{2-6}\cmidrule{8-12}
{\fexp Exp}-Ratio 				&0.388 {\frank (3)} 		&0.400 {\frank (5)} 		&0.415 {\frank (10)} 	&0.440 {\frank (8)} 		&	0.420 {\frank (4)} 		&&0.613 {\frank (4)} 	&	0.604 {\frank (5)} 	&0.587 {\frank (9)} 		&0.556 {\frank (8)}		&0.576 {\frank (4)}		&{\frank (6.0)}\\
{\fmed Med}-Ratio 				&0.450 {\frank (16)} 	&0.418 {\frank (11)} 	&0.388 {\frank (1)*}	&0.428 {\frank (6)} 		&	0.410 {\frank (2)} 		&&0.554 {\frank (15)} &	0.586 {\frank (12)}	&0.619 {\frank (1)*}	&0.571 {\frank (5)}		&0.579 {\frank (3)}		&{\frank (7.2)}\\
{\fmod Mod}-Ratio 				&0.400 {\frank (7)} 		&0.370 {\frank (1)*}	&0.408 {\frank (5)} 		&0.435 {\frank (7)} 		&	0.428 {\frank (5)} 		&&0.595 {\frank (8)} 	&	0.634 {\frank (1)*}	&0.591 {\frank (7)} 		&0.558 {\frank (7)}		&0.560 {\frank (9)}		&{\frank (5.7)}\\
{\fphi $\varphi$Pr}-Ratio	&0.372 {\frank (1)*}	&0.430 {\frank (12)} 	&0.410 {\frank (7)} 		&0.415 {\frank (2)} 		&	0.408 {\frank (1)*} 		&&0.630 {\frank (1)*} &	0.589 {\frank (9)} 	&0.590 {\frank (8)} 		&0.591 {\frank (2)}		&0.589 {\frank (1)*}	&{\frank (4.4)}\\
{\fbaseline Ratio} 		&0.515 {\frank (20)} 		&	0.520 {\frank (21)} 		&	0.490 {\frank (18)} 	&	0.475	 {\frank (17)} 		&	0.498	 {\frank (19)} 		&&0.550 {\frank (16)} 	&	0.461 {\frank (22)} 	&0.461 {\frank (21)} 		&	0.503 {\frank (19)}		&	0.517 {\frank (20)}		&{\frank (19.3)}\\
\cmidrule{2-6}\cmidrule{8-12}
{\fexp Exp}-Gtest 				&0.393 {\frank (6)} 		&0.403 {\frank (7)} 		&0.413 {\frank (8)} 		&0.420 {\frank (3)} 		&	0.435 {\frank (9)} 		&&0.610 {\frank (5)} 	&	0.588 {\frank (10)}	&0.582 {\frank (10)} 	&0.586 {\frank (3)}		&0.563 {\frank (5)}		&{\frank (6.6)}\\
{\fmed Med}-Gtest 				&0.437 {\frank (13)} 	&0.400 {\frank (5)} 		&0.408 {\frank (5)} 		&0.420 {\frank (3)} 		&	0.453 {\frank (15)} 		&&0.559 {\frank (14)} &	0.590 {\frank (8)} 	&0.600 {\frank (2)} 		&0.580 {\frank (4)}		&0.551 {\frank (10)}	&{\frank (7.9)}\\
{\fmod Mod}-Gtest 				&0.448 {\frank (15)} 	&0.383 {\frank (4)} 		&0.398 {\frank (2)} 		&0.428 {\frank (5)} 		&	0.433 {\frank (6)} 		&&0.571 {\frank (12)} &	0.622 {\frank (4)} 	&0.593 {\frank (5)} 		&0.565 {\frank (6)}		&0.551 {\frank (10)}	&{\frank (6.9)}\\
{\fphi $\varphi$Pr}-Gtest	&0.450 {\frank (16)} 	&0.445 {\frank (14)} 	&	0.443{\frank (12)}	&0.455 {\frank (11)} 	&	0.433 {\frank (6)} 		&&0.544 {\frank (19)} &	0.555 {\frank (16)}	&0.552 {\frank (12)} 	&0.538 {\frank (13)}	&0.562 {\frank (7)}		&{\frank (12.6)}\\
{\fbaseline Gtest} 				&0.440 {\frank (14)} 		&	0.505 {\frank (19)} 		&	0.501 {\frank (21)} 		&	0.486 {\frank (19)} 	&	0.471 {\frank (16)} 		&&	0.542 {\frank (20)} 	&		0.492 {\frank (19)}	&	0.490 {\frank (20)} 	&	0.499 {\frank (21)}		&	0.534 {\frank (16)}		&{\frank (18.5)}\\
\cmidrule{2-6}\cmidrule{8-12}
{\fexp Exp}-Conf 				&0.405 {\frank (8)} 		&0.415 {\frank (10)} 	&0.453 {\frank (13)} 	&0.455 {\frank (11)} 	&	0.448 {\frank (12)} 		&&0.595 {\frank (8)} 	&	0.587 {\frank (11)}	&0.539 {\frank (15)} 	&0.543 {\frank (12)}	&0.535 {\frank (15)}	&{\frank (11.5)}\\
{\fmed Med}-Conf 				&0.378 {\frank (2)} 		&0.373 {\frank (2)} 		&0.438 {\frank (11)} 	&0.463 {\frank (15)} 	&	0.435 {\frank (9)} 		&&0.629 {\frank (2)} 	&	0.632 {\frank (2)} 	&0.555 {\frank (11)} 	&0.536 {\frank (15)}	&0.545 {\frank (13)}	&{\frank (8.2)}\\
{\fmod Mod}-Conf 				&0.392 {\frank (5)} 		&0.373 {\frank (2)} 		&0.400 {\frank (3)} 		&0.440 {\frank (8)} 		&	0.435 {\frank (9)} 		&&0.606 {\frank (6)} 	&	0.627 {\frank (3)} 	&0.600 {\frank (2)} 		&0.556 {\frank (8)}		&0.563 {\frank (5)}		&{\frank (5.1)}\\
{\fphi $\varphi$Pr}-Conf	&0.468 {\frank (19)} 	&0.460 {\frank (15)} 	&	0.495 {\frank (20)} 	&0.505 {\frank (21)} 	&	0.485 {\frank (17)} 		&&0.547 {\frank (18)} &	0.556 {\frank (15)}	&0.519 {\frank (16)} 	&0.507 {\frank (18)}	&0.540 {\frank (14)}	&{\frank (17.3)}\\
{\fbaseline Conf} 	&0.455 {\frank (18)} 		&	0.500 {\frank (18)} 	&	0.460 {\frank (15)} 	&	0.464 {\frank (16)} 	&		0.450 {\frank (14)} &&	0.514 {\frank (21)} 	&	0.479 {\frank (20)}	&	0.510 {\frank (17)} 	&	0.498 {\frank (22)}	&	0.519 {\frank (19)}	&{\frank (18.0)}\\
\cmidrule{2-6}\cmidrule{8-12}
Exp-Freq	&0.423 {\frank (10)}	&0.465 {\frank (16)} 	&0.508 {\frank (22)} 	&0.498 {\frank (20)} 	&	0.505 {\frank (22)} 		&&0.579 {\frank (11)} &	0.549 {\frank (17)}	&0.496 {\frank (19)} 	&0.513 {\frank (16)}	&	0.498 {\frank (22)}	&{\frank (17.5)}\\
{\fbaseline Freq}		&0.515 {\frank (20)}	&	0.520 {\frank (21)} 	&	0.490 {\frank (18)} 	&	0.475 {\frank (17)} 	&		0.498 {\frank (19)} 		&&	0.550 {\frank (16)} &		0.461 {\frank (21)}	&	0.461 {\frank (21)} 	&	0.503 {\frank (19)}	&	0.517 {\frank (20)}	&{\frank (19.2)}\\
\bottomrule
\end{tabular}
}
\end{table*}

\begin{table*}[t]
\caption{Results on the HIV (Human Immunodeficiency Virus) dataset with different number of features ($t=100, \cdots, 500$). The results are reported as ``average performance + (rank)". }\label{tab:hiv}
{\tiny
\centering
\begin{tabular}{rlllllrlllllr}
\toprule
 & \multicolumn{5}{c}{Error Rate $\downarrow$}&&\multicolumn{5}{c}{F1 $\uparrow$}&Avg.\\
\cmidrule{2-6}\cmidrule{8-12}
{Methods}&\multicolumn{1}{c}{$t=100$}&\multicolumn{1}{c}{$t=200$}&\multicolumn{1}{c}{$t=300$}&\multicolumn{1}{c}{$t=400$}&\multicolumn{1}{c}{$t=500$}&&\multicolumn{1}{c}{$t=100$}&\multicolumn{1}{c}{$t=200$}&\multicolumn{1}{c}{$t=300$}&\multicolumn{1}{c}{$t=400$}&\multicolumn{1}{c}{$t=500$}&\multicolumn{1}{c}{Rank}\\
\midrule
{\fexp Exp}-HSIC 				&0.480 {\frank (15)}	&0.470 {\frank (10)} 	&0.489 {\frank (12)} 	&0.505 {\frank (16)} 	&	0.498 {\frank (13)} 		&&0.526 {\frank (13)} &	0.531 {\frank (8)}	&0.517 {\frank (11)} 	&0.491 {\frank (14)}	&	0.492 {\frank (13)}	&{\frank (12.5)}\\
{\fmed Med}-HSIC  				&0.498 {\frank (17)} 		&0.500 {\frank (18)} 		&0.470 {\frank (7)} 		&0.484 {\frank (11)} 	&	0.507 {\frank (16)} 		&&0.501 {\frank (18)} &	0.493 {\frank (18)} 	&0.526 {\frank (8)} 		&0.510 {\frank (10)}	& 0.474 {\frank (16)}		&{\frank (13.9)}\\
{\fmod Mod}-HSIC 				&0.502 {\frank (18)} 		&0.489 {\frank (15)} 		&	0.482 {\frank (11)} 	&0.498 {\frank (14)}	&0.500 {\frank (14)} 			&&0.501 {\frank (18)} 	&	0.501 {\frank (16)} 	&0.495 {\frank (14)} 		&	0.481 {\frank (17)}	&	0.467 {\frank (19)}		&{\frank (15.6)}\\
{\fphi $\varphi$Pr}-HSIC	&0.523 {\frank (19)} 	&0.511 {\frank (19)} 	&0.516 {\frank (18)} 	&0.525 {\frank (19)} 	&	0.523 {\frank (20)} 		&&	0.484 {\frank (20)} 	&		0.492 {\frank (19)}	&	0.481 {\frank (16)} 	&	0.474 {\frank (19)}	&	0.482 {\frank (14)}	&{\frank (18.3)}\\
{\fbaseline HSIC}			&0.464 {\frank (6)}	&	0.495 {\frank (17)} 	&	0.566 {\frank (21)} 	& 0.500 {\frank (15)} 	&0.505	{\frank (15)} 		&&0.526	 {\frank (13)} &	0.460 {\frank (20)}	&	0.405 {\frank (21)} 	&0.489 {\frank (15)}	&	0.471 {\frank (18)}	&{\frank (16.1)}\\
\cmidrule{2-6}\cmidrule{8-12}
{\fexp Exp}-Ratio 				&0.475 {\frank (13)} 		&	0.477 {\frank (11)} 		&	0.491 {\frank (13)} 	&	0.516 {\frank (18)} 		&		0.484 {\frank (8)} 		&&	0.541 {\frank (8)} 	&		0.533 {\frank (7)} 	&	0.509 {\frank (13)} 		&	0.477 {\frank (18)}		&	0.519 {\frank (8)}		&{\frank (11.3)}\\
{\fmed Med}-Ratio 				&0.466 {\frank (8)} 	&	0.464 {\frank (8)} 	&	0.470 {\frank (7)}	&	0.457 {\frank (5)} 		&	0.473 {\frank (6)} 		&&	0.541 {\frank (8)} &		0.528 {\frank (9)}	&	0.524 {\frank (9)}	&	0.534 {\frank (6)}		&	0.521 {\frank (6)}		&{\frank (7.2)}\\
{\fmod Mod}-Ratio 				&0.450 {\frank (3)} 		&	0.452 {\frank (5)}	&	0.466 {\frank (4)} 		&	0.480 {\frank (9)} 		&		0.484 {\frank (8)} 		&&	0.558 {\frank (5)} 	&		0.547 {\frank (5)}	&	0.528 {\frank (6)} 		&	0.509 {\frank (11)}		&	0.500 {\frank (12)}		&{\frank (6.8)}\\
{\fphi $\varphi$Pr}-Ratio	&0.473 {\frank (11)}	&	0.480 {\frank (12)} 	&	0.466 {\frank (4)} 		&	0.470 {\frank (8)} 		&		0.468 {\frank (5)} 		&&	0.544 {\frank (7)} &		0.519 {\frank (13)} 	&	0.538 {\frank (5)} 		&	0.531 {\frank (7)}		&	0.538  {\frank (5)}	&{\frank (7.7)}\\
{\fbaseline Ratio}			&0.530 {\frank (21)} 		&	0.486 {\frank (13)} 		&	0.589 {\frank (22)} 	&	0.411 {\frank (1)*} 		&		0.520 {\frank (19)} 		&&	0.456 {\frank (21)} 	&	0.495	 {\frank (17)} 	&	0.376 {\frank (22)} 		&	0.562 {\frank (4)}		&	0.443 {\frank (20)}		&{\frank (16)}\\
\cmidrule{2-6}\cmidrule{8-12}
{\fexp Exp}-Gtest 				&0.468 {\frank (9)} 		&	0.466 {\frank (9)} 		&	0.468 {\frank (6)} 		&	0.466 {\frank (7)} 		&		0.482 {\frank (7)} 		&&	0.562 {\frank (4)} 	&		0.565 {\frank (4)}	&	0.548 {\frank (4)} 	&	0.537 {\frank (5)}		&	0.520 {\frank (7)}		&{\frank (6.2)}\\
{\fmed Med}-Gtest 				&0.464 {\frank (6)} 	&	0.461 {\frank (7)} 		&	0.507 {\frank (17)} 		&	0.507 {\frank (17)} 		&		0.511 {\frank (17)} 		&&	0.534 {\frank (11)} &		0.520 {\frank (11)} 	&	0.480 {\frank (17)} 		&	0.483 {\frank (16)}		&	0.474 {\frank (16)}	&{\frank (10.9)}\\
{\fmod Mod}-Gtest 				&0.477 {\frank (14)} 	&	0.486 {\frank (13)} 		&	0.475 {\frank (10)} 		&	0.491 {\frank (13)} 		&		0.489 {\frank (11)} 		&&	0.529 {\frank (12)} &		0.507 {\frank (14)} 	&	0.523 {\frank (10)} 		&	0.497 {\frank (13)}		&	0.501 {\frank (11)}	&{\frank (12.1)}\\
{\fphi $\varphi$Pr}-Gtest	&0.430	 {\frank (1)*} 	&0.420	 {\frank (2)} 	&	0.425 {\frank (1)*}	&	0.418 {\frank (2)} 	&		0.425 {\frank (2)} 		&&	0.617 {\frank (1)*} &		0.633 {\frank (1)*}	&	0.630 {\frank (1)*} 	&	0.637 {\frank (1)*}	&	0.633 {\frank (1)*}		&{\frank *(1.3)}\\
{\fbaseline Gtest} 			& 0.473 {\frank (11)} 		&		0.550 {\frank (21)} 		&	0.493 {\frank (14)} 		& 0.534 {\frank (20)} 		& 0.493 {\frank (12)} &&	0.514 {\frank (16)} 	&	0.426 {\frank (22)}	&		0.491 {\frank (15)} 	&	0.509 {\frank (11)}		&	0.477 {\frank (15)}		&{\frank (15.7)}\\
\cmidrule{2-6}\cmidrule{8-12}
{\fexp Exp}-Conf 				&0.457 {\frank (4)} 		&	0.430 {\frank (4)} 	&	0.441 {\frank (2)} 	&	0.443 {\frank (4)} 	&		0.441 {\frank (3)} 		&&	0.576 {\frank (3)} 	&		0.590 {\frank (2)}	&	0.572 {\frank (2)} 	&	0.570 {\frank (3)}	&	0.573 {\frank (4)}	&{\frank (3.1)}\\
{\fmed Med}-Conf 				&0.445 {\frank (2)} 		&	0.427 {\frank (3)} 		&	0.441 {\frank (2)} 	&	0.441 {\frank (3)} 	&		0.443 {\frank (4)} 		&&	0.579 {\frank (2)} 	&		0.588 {\frank (3)} 	&	0.572 {\frank (2)} 	&	0.579 {\frank (2)}	&	0.574 {\frank (3)}	&{\frank (2.6)}\\
{\fmod Mod}-Conf 				&0.457 {\frank (4)} 		&	0.455 {\frank (6)} 		&	0.473 {\frank (9)} 		&	0.482 {\frank (10)} 		&		0.484 {\frank (8)} 		&&	0.556 {\frank (6)} 	&		0.545 {\frank (6)} 	&	0.527 {\frank (7)} 		&	0.518 {\frank (9)}		&	0.508 {\frank (9)}		&{\frank (7.4)}\\
{\fphi $\varphi$Pr}-Conf	&0.534 {\frank (22)} 	&	0.552 {\frank (22)} 	&		0.545 {\frank (19)} 	&	0.548 {\frank (21)} 	&		0.541 {\frank (22)} 		&&	0.454 {\frank (22)} &		0.443 {\frank (21)}	&	0.444 {\frank (20)} 	&	0.443 {\frank (22)}	&	0.438 {\frank (21)}	&{\frank (21.2)}\\
{\fbaseline Conf} 	&0.468	 {\frank (9)} 		&	0.416	 {\frank (1)*} 	&	0.502 {\frank (15)} 	&	0.489 {\frank (12)} 	&		0.339 {\frank (1)*} 		&&	0.515 {\frank (15)} 	&	0.528 {\frank (9)}	&	0.468 {\frank (19)} 	&	0.462 {\frank (20)}	&	0.621 {\frank (2)}	&{\frank (10.3)}\\
\cmidrule{2-6}\cmidrule{8-12}
Exp-Freq				&0.525 {\frank (20)}	&	0.520 {\frank (20)} 	&	0.548 {\frank (20)} 	&	0.550 {\frank (22)} 	&		0.527 {\frank (21)} 		&&	0.503 {\frank (17)} &		0.520 {\frank (11)}	&	0.473 {\frank (18)} 	&	0.457 {\frank (21)}	&		0.423 {\frank (22)}	&{\frank (19.2)}\\
{\fbaseline Freq}				&0.489 {\frank (16)}	&	0.489 {\frank (15)} 	&	0.502 {\frank (15)} 	&	0.461 {\frank (6)} 	&		0.514 {\frank (18)} 		&&	0.535 {\frank (10)} &		0.505 {\frank (15)}	&	0.517 {\frank (11)} 	&	0.520 {\frank (8)}	&		0.502 {\frank (10)}	&{\frank (12.4)}\\
\bottomrule
\end{tabular}
}
\end{table*}

\subsection{Comparative Methods}
We compared our method using different statistical measures and discrimination score functions summarized as follows:

\noindent\textbullet\  \textit{Frequent Subgraphs + Expectation (Exp+Freq)}: The first baseline method is finding frequent subgraph features within uncertain graphs. This baseline is similar to the method introduced in \cite{ZLGZ09}. In our data model,  this baseline method computes the exact expected frequency of each subgraph features, instead of approximated values. The top ranked frequent patterns are extracted as used as features for graph classification.

\noindent\textbullet\  \textit{{\guc} with HSIC based discrimination scores}: we compare with four different versions of our {\guc} method based upon HSIC criterion, which maximize the dependence between subgraph features and graph labels \cite{KFY11}. {``{\fexp Exp}-HSIC"} computes the expected HSIC value for each subgraph feature, and find the top-$k$ subgraphs with the largest values. {``{\fmed Med}-HSIC"} computes the median HSIC value for each subgraph feature, while {``{\fmod Mod}-HSIC"} computes the mode HSIC value. {``{\fphi $\varphi\Pr$}-HSIC"} computes the $\varphi$-probability of HSIC value for each subgraph feature.

\noindent\textbullet\  \textit{{\guc} with Frequency Ratio based discrimination scores}: we also compare our method based upon Frequency Ratio, {\ie}, {``{\fexp Exp}-Ratio"}, {``{\fmed Med}-Ratio"}, {``{\fmod Mod}-Ratio"} and {``{\fphi $\varphi\Pr$}-Ratio".

\noindent\textbullet\  \textit{{\guc} with G-test based discrimination scores}: we then compare our method based upon G-test criterion, {\ie}, {``{\fexp Exp}-Gtest"}, {``{\fmed Med}-Gtest"}, {``{\fmod Mod}-Gtest"} and ``{\fphi $\varphi\Pr$}-Gtest"}.

\noindent\textbullet\  \textit{{\guc} with Confidence based discrimination scores}: the 5th group of methods are based upon G-test criterion, {\ie}, {``{\fexp Exp}-Conf"}, {``{\fmed Med}-Conf"}, {``{\fmod Mod}-Conf"} and {``{\fphi $\varphi\Pr$}-Conf"}.

\noindent\textbullet\  \textit{Simple Thresholding}: Another group methods we have compared are the feature selection methods for certain graphs. In order to get the certain graphs from the uncertain graphs in the dataset, we perform simple tresholding over the weights of the links to get the binary links. These baseline methods include: ``{\fbaseline Freq}'', ``{\fbaseline HISC}", ``{\fbaseline Ratio}", ``{\fbaseline Gtest}" and ``{\fbaseline Conf}", which correspond to the discrimination scores used in previous 5 groups separately.

LibSVM \cite{libsvm} with the linear kernel is used as the base
classifier for all compared methods. The $min\_sup$ in the gSpan for ADHD, ADNI and HIV datasets are $20\%$, $40\%$ and $40\%$ respectively. Since the range of different discrimination functions can be extremely different. We set the default $\varphi$ for HSIC criterion, G-test score, frequency ratio and confidence as 0.03, 200, 1 and 0.5, respectively.

\subsection{Performance on  Uncertain Graph Classification}
In our experiments, we first randomly sample 80\% of the uncertain graphs as the training set, and the remaining graphs as the test set. This random sampling experiment was repeated 20 times. The average performances with the rank of each method are reported. The reason for using classification performances to evaluate the quality of subgraph features is that classification methods can usually achieve higher accuracy with features of better discriminative powers. We measure the classification performance by error rate and F1 score.

Table~\ref{tab:adhd} and Table~\ref{tab:adni} show the evaluation results in terms of classification error rates and F1 scores with different number of selected subgraph features ($t=100, \cdots, 500$). The results of each method are shown with average performance values and their ranks among all the other methods. Values with $*$ stand for the best performance for the corresponding evaluation criterion. It is worth noting that the neuroimaging tasks are generally very hard to predict very accurately. According to a global competition on ADHD dataset\footnote{http://www.childmind.org/en/posts/press-releases/2011-10-12-johns-hopkins-team-wins-adhd-200-competition}, the average performance of all winning teams is about 8\% over the prediction accuracy of chance (i.e., randomly assigning diagnoses). Thus in neuroimaging tasks, it is very hard for classification algorithms to achieve even moderate error rates. And in ADHD dataset, the best performance that {\guc} can achieve is with error rate 37\%, which is 13\% improvement over the prediction error rate of chance.

We find that our discriminative subgraph mining method with different settings outperforms the baseline method (Exp-Freq) for frequent subgraph mining, which selects subgraph features based upon expected frequencies in the uncertain graph dataset. This is because that frequent subgraph features in uncertain graph dataset may not be relevant to the classification task. 

Moreover, we can see that almost all the {\guc} methods outperform the simple thresholding methods which directly convert the uncertain graphs into certain graphs and then use different discimination functions to select subgraph features. This is because that simply converting uncertain graphs into certain graph can loss the uncertainty information about the linkage structures of the graphs, thus the classification performances on certain graphs are not as good as the performance of uncertain graphs.

A third observation is that the performance of each method on different dataset can be quite different. However, the best methods that  consistently outperforms other methods in all datasets are Med-Conf and $\varphi$-Pr-Ratio. They both have their advantages in different perspectives.  Med-Conf method has one less parameter than that of  $\varphi$-Pr-Ratio. $\varphi$-Pr-Ratio method has an additional subgraph pruning bound compared to  Med-Conf method, which can be important for datasets with even larger graphs.

\subsection{Influence of Parameter}

\begin{figure}[t]
  \centering
  \subfigure[$\varphi$ vs Error Rate]{\label{subfig:adni_phi_er}
    \begin{minipage}[l]{0.68\columnwidth}
      \centering
      \includegraphics[width=1\textwidth]{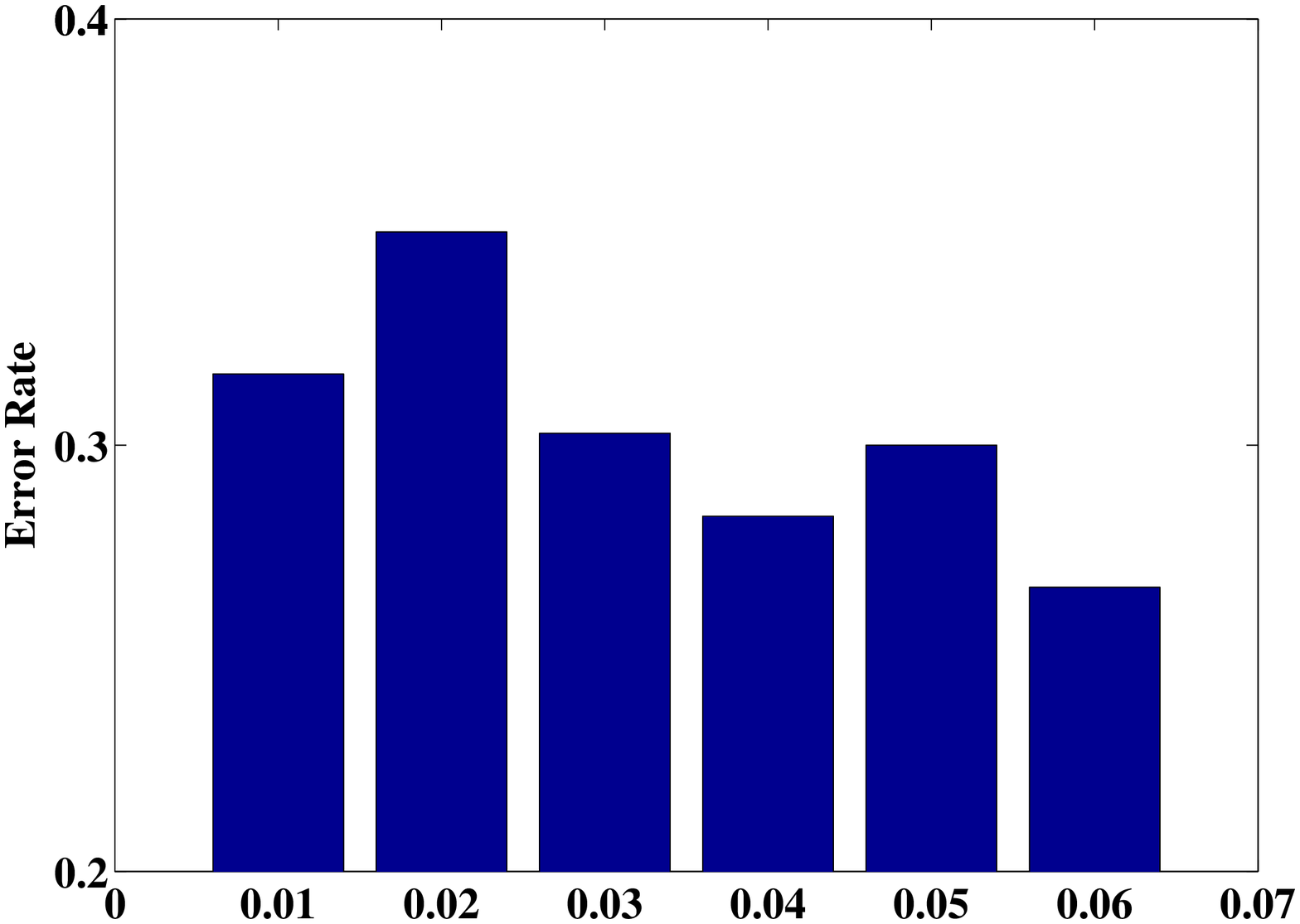}
    \end{minipage}
  }
    \subfigure[$\varphi$ vs F1]{\label{subfig:adni_phi_f1}
    \begin{minipage}[l]{0.68\columnwidth}
      \centering
      \includegraphics[width=1\textwidth]{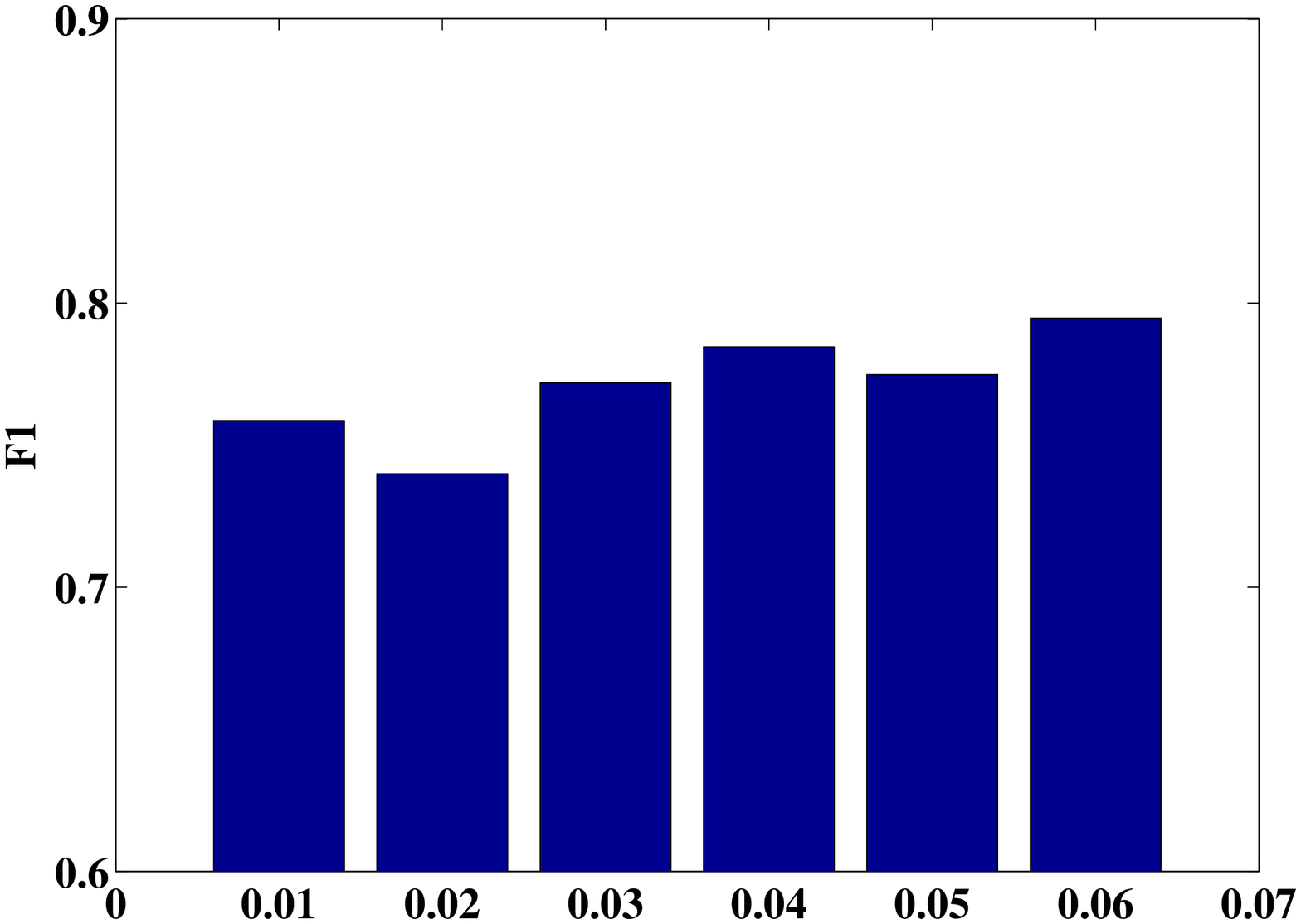}
    \end{minipage}
  } 
  \subfigure[ Pruning]{\label{subfig:adni_bound}  
    \begin{minipage}[l]{0.62\columnwidth}
      \centering
      \includegraphics[width=1\textwidth]{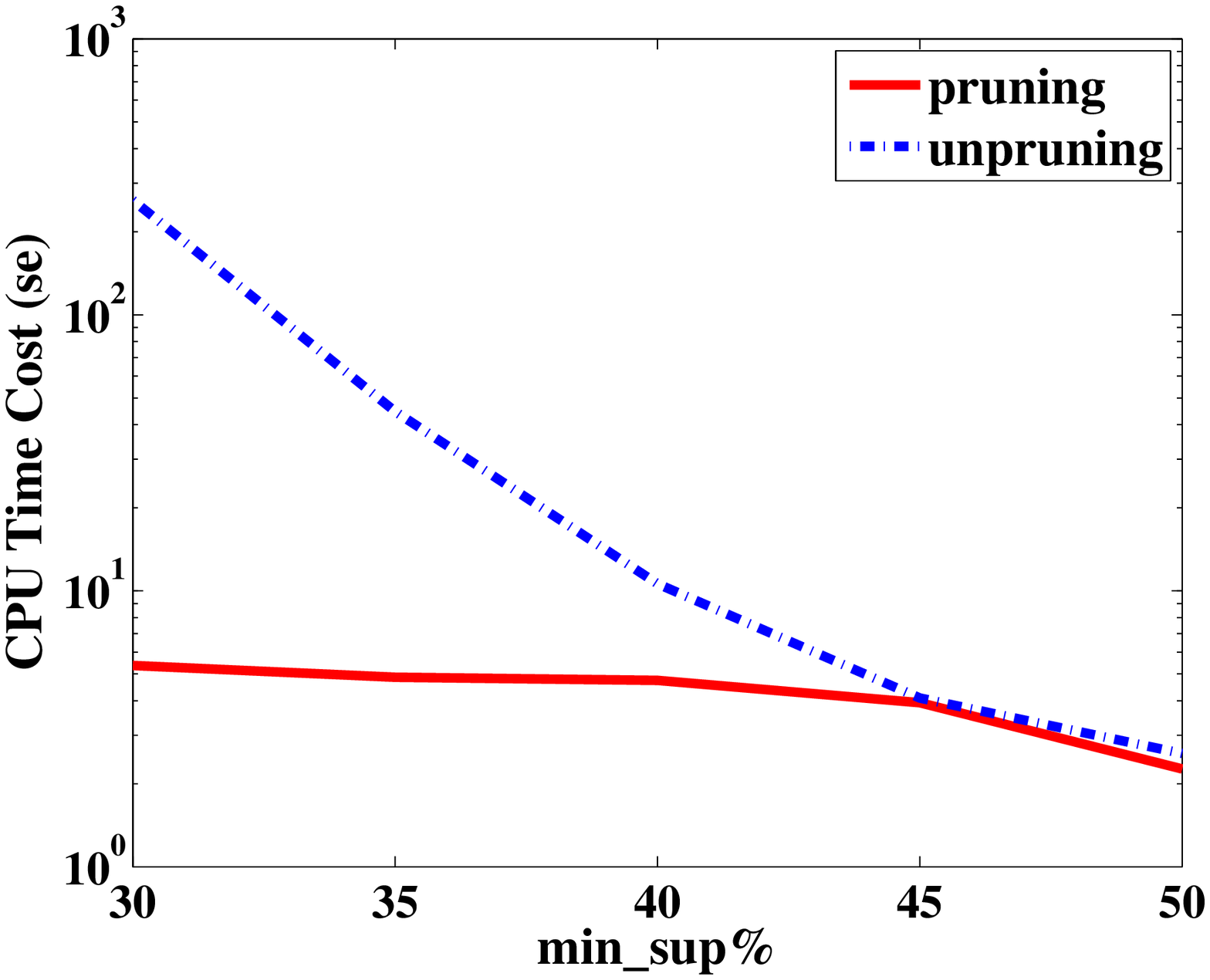}
    \end{minipage}
  }
  \caption{Parameter Studies  (ADNI dataset).} \label{fig:exp2}
\end{figure}

In the $\varphi$-Pr based methods, there is an additional threshold parameter than the other methods. In Figure~\ref{subfig:adni_phi_er} and Figure~\ref{subfig:adni_phi_f1}, we tested the $\varphi$-Pr-HSIC with $\varphi$ values among $\{0.01, 0.02, \cdots 0.06\}$ separately. We can see that the method is not  sensitive to the parameter $\varphi$. Generally,  the performance of $\varphi$-Pr-HSIC with default setting ($\varphi =0.03$) is pretty good. If we try to optimize the selection of $\varphi$ value, the accuracy can be even better.

\begin{figure}[t]
 \centering
 \begin{minipage}[l]{0.64\columnwidth}
\includegraphics[width=1\textwidth]{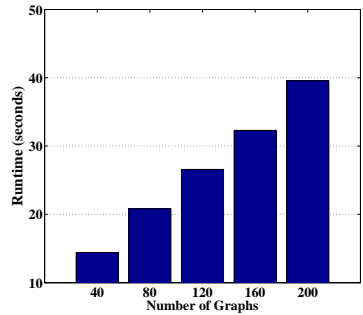}
    \end{minipage}
\caption{Running time on ADNI dataset.} \label{subfig:result_num_graphs}
\end{figure}

We also compare {\guc} models with and without pruning in the subgraph search space as summarized in Figure~\ref{subfig:adni_bound}. The CPU time with different $min\_sup$ for Exp-HSIC in ADNI dataset is reported.  {\guc} can improve the efficiencies by pruning the subgraph search space. In other datasets {\guc} shows similar trends. Figure~\ref{subfig:result_num_graphs} shows the running time for mod-HISC with different number of graphs in the dataset. In addition to the dynamic programming method we used in {\guc}, we also find that the brute-force searching method that enumerates all possible worlds of the uncertain graph dataset cannot work on small datasets with even 40 graphs.
The running time of {\guc} scales almost linearly with the number of graphs in the dataset. Althought the dynamic programming process of {\guc} is $O(n^2)$, which is quadratic to the number of graphs in the dataset. However, in the ADHD dataset, the main computational cost of {\guc} algorithm is for the subgraph enumeration step, which is linear to the number of the graphs in the dataset. In cases of even larger datasets, the dynamic programming process could eventually dominant the computational cost. In these cases, the divide-and-conquer method  in \cite{SCCC10} could be used to further optimize the computational cost. 

%


\section{Related Work} \label{sec:relatedwork}
Our work is related to subgraph mining techniques for both certain graphs and uncertain graphs. We briefly discuss both of them.

Mining subgraph features in graph data has been studied intensively in recent years \cite{KK01}. Most of the previous research has been focused on certain graph datasets, where the edges of the graph objects are binary/certain. The aim of these subgraph mining method is to extract important subgraph features based on the structure of the graphs. Depending on whether the class labels are considered in the feature mining steps, existing methods can roughly be categorized into two types: frequent subgraph mining and discriminative subgraph mining.  In frequent subgraph mining, Yan and Han proposed a depth-first search algorithm, gSpan \cite{YH02}, which maps each graph to a unique minimum DFS code and use right-most extension technique for subgraph extension. There are also many other frequent subgraph mining methods that have been proposed, {\eg}, AGM \cite{IWM00}, FSG \cite{KK01}, MoFa \cite{BB02}, and Gaston \cite{NK04}, etc. Discriminative subgraph mining have also been studied intensively in the literature, such as LEAP \cite{YCHY08} and LTS \cite{JW11}, where the task is to find discriminative subgraph for graph classifications. 

Recently, there has been a growing interest in exploiting data uncertainty, especially structural uncertainty in graph data. There are some recent works on mining frequent subgraph features for uncertain graphs \cite{ZLGZ09,ZGL10,ZLGZ10,PIS11}. The problem of mining frequent subgraph in uncertain graphs are more difficult to those of certain graphs. The authors \cite{ZLGZ09} proposed a method to estimate approximately the expected support of a subgraph feature in uncertain graph datasets. In \cite{ZGL10}, the authors studies the $\varphi$-probabilities for frequent subgraph features within uncertain graph datasets. The difference between these works and our paper are as follows: 1) In this paper, we study how to find discriminative subgraph features for uncertain graph data. The class labels of the graph objects are considered during the subgraph mining. 2) The graph model in our paper is different from previous uncertain graph data, since we assume different graph object shares the same set of nodes as inspired by the neuroimaging applications. Thus, our method compute the exact discrimination scores of each subgraph features, instead of approximate scores. There are also many other works on uncertain graphs, which focus on different problems, {\eg}, reliable subgraph mining \cite{JLA11}, $k$-nearest neighbor discovery \cite{PBGK10}, subgraph retrieval \cite{YWWC11} {\etc}

Our work is also motivated by the recent advances in analyzing neuroimaging data using data mining and  machine learning approaches \cite{HLS09,HSY10,HLY11,ZYLY11}.  Huang et. al. \cite{HLS09} developed a sparse inverse covariance estimation method for analyzing brain connectivities in PET images of patients with Alzheimer's disease. 



\section{Conclusion}\label{sec:conclusion}
In this paper, we studied the problem of discriminative subgraph feature selection for uncertain graph classification. We proposed a general framework, called {\guc}, for finding discriminative subgraph feature in uncertain graphs based upon various statistical measures. The probability distributions of the scoring function are efficiently computed based on dynamic programming.

\section*{Acknowledgment}
This work is supported in part by NSF through grants IIS-0905215,
CNS-1115234, IIS-0914934, DBI-0960443, and OISE-1129076, 
Huawei and KAU grants, NIH R01-MH080636 grant.

\balance
\bibliographystyle{abbrv}
\bibliography{reference}

\end{document}